\documentclass[final]{cvpr}
\usepackage{times}
\usepackage{epsfig}
\usepackage{graphicx}
\usepackage{amsmath}
\usepackage{amssymb}

\usepackage{microtype}
\usepackage{subfig}
\usepackage{booktabs}
\usepackage{caption}
\usepackage{pifont}
\usepackage[dvipsnames]{xcolor}
\usepackage[hyphens]{url}
\usepackage{float}
\usepackage{algorithm}
\usepackage{algorithmic}

\usepackage[pagebackref=true,breaklinks=true,colorlinks,bookmarks=false]{hyperref}

\def\cvprPaperID{7} 
\def\confYear{CVPR 2021}

\begin{document}

\title{Training Dynamical Binary Neural Networks with Equilibrium Propagation}


\author{Jérémie Laydevant$^{1*}$, Maxence Ernoult$^{2 \dag}$, Damien Querlioz$^{3}$, Julie Grollier$^{1}$\\
$^{1}$ Unité mixte de Physique, CNRS-Thales, Université Paris-Saclay, Palaiseau, France\\
$^{2}$ MILA, Montreal, Canada\\
$^{3}$ Université Paris-Saclay, CNRS, Centre  de  Nanosciences  et  de  Nanotechnologies, Palaiseau, France\\
{\tt\small \{$^{*}$jeremie.laydevant@cnrs-thales.fr}; 
{$^{\dag}$\tt\small ernoultm@mila.quebec\}}
}

\maketitle
\thispagestyle{empty}

\begin{abstract}
   Equilibrium Propagation (EP) is an algorithm intrinsically adapted to the training of physical networks,
    thanks to the local updates of weights given by the internal dynamics of the system. However, the construction of such a hardware requires to make the algorithm compatible with existing neuromorphic CMOS technologies, which generally exploit digital communication between neurons and offer a limited amount of local memory. In this work, we demonstrate that EP can train dynamical networks with binary activations and weights. We first train systems with binary weights and full-precision activations, achieving an accuracy equivalent to that of full-precision models trained by standard EP on MNIST, and losing only 1.9\% accuracy on CIFAR-10 with equal architecture. We then extend our method to the training of models with binary activations and weights on MNIST, achieving an accuracy within 1\% of the full-precision reference for fully connected architectures and reaching the full-precision accuracy for convolutional architectures. Our extension of EP to binary networks opens new solutions for on-chip learning and provides a compact framework for training BNNs end-to-end with the same circuitry as for inference.
\end{abstract}


\vspace*{-0.15in}
\section{Introduction}
\label{introduction}

Conventional deep learning models, trained with error backpropagation (BP), have demonstrated outstanding performance at multiple cognitive tasks. But the training process is so energy consuming \cite{greenAI_schwartz_2020, strubell2019} that it questions the environmental sustainability of AI deployment \cite{lacoste2019quantifying}. These artificial neural networks are actually trained on un-optimized von Neumann hardware with a delocalized memory, such as GPUs or TPUs. Furthermore, they struggle to fully benefit from a hardware which provides local memory access at neuron level as their learning rule, backpropagation, is fundamentally non-local.

Equilibrium Propagation (EP) \cite{DBLP:journals/corr/ScellierB16} is a learning framework that leverages the dynamics of energy-based physical systems fed by static inputs to compute weight updates with a learning rule local in space \cite{NIPS2019_8930} which can also be made local in time \cite{ernoult2020equilibrium}, and in addition scales to CIFAR-10 \cite{laborieux2020scaling}. Today, EP is developed on standard hardware (GPU) that 1) does not provide for the low power and the computing efficiency a dedicated hardware implementation might exhibit \cite{thakur_large-scale_2018, markovic_physics_2020} and 2) prevents EP to scale to large scale datasets such as ImageNet due to the duration of simulations. An EP-dedicated hardware would reduce the energy consumption of training by two orders of magnitude compared to GPUs and accelerate training by several orders of magnitude  \cite{martin_eqspike_2021}, while being competitive on large scale benchmarks in terms of accuracy since the gradients estimates prescribed by EP are equivalent to those given by BPTT \cite{NIPS2019_8930}.

The main asset of EP is the ability of on-chip learning, especially when the memory and the computational budgets dedicated to training and inference are constrained (e.g. embedded environments). EP also naturally suits for training physical systems intrinsically dynamical whose dynamics are unknown and hardly derivable \cite{kendall_training_2020,zoppo2020equilibrium,martin_eqspike_2021}. EP therefore appears as a solution for on-chip training for embedded systems and dynamical hardware, two cases with which BP is not compatible without major adaptation.

EP is however based on full-precision (64 bits floating point) weights and activations that do not match the current requirements of such hardware systems. Full-precision weights overload the memory capacity of chips when they are stored digitally \cite{gopalakrishnan2020,thakur_large-scale_2018}, and are prone to noise and hard to read when stored with emerging synaptic nano-devices \cite{ambrogio2018, zhang_neuro-inspired_2020, hirtzlin2019hybrid, hirtzlin2020}. Moreover, analog activation functions are not directly compatible with widely-used digital communications between neurons \cite{thakur_large-scale_2018}. 

\begin{figure}[h]
  \centering
  \subfloat[a][Binary weights (Section \ref{sec:bin-W-eqprop})]{\includegraphics[width=0.5\textwidth]{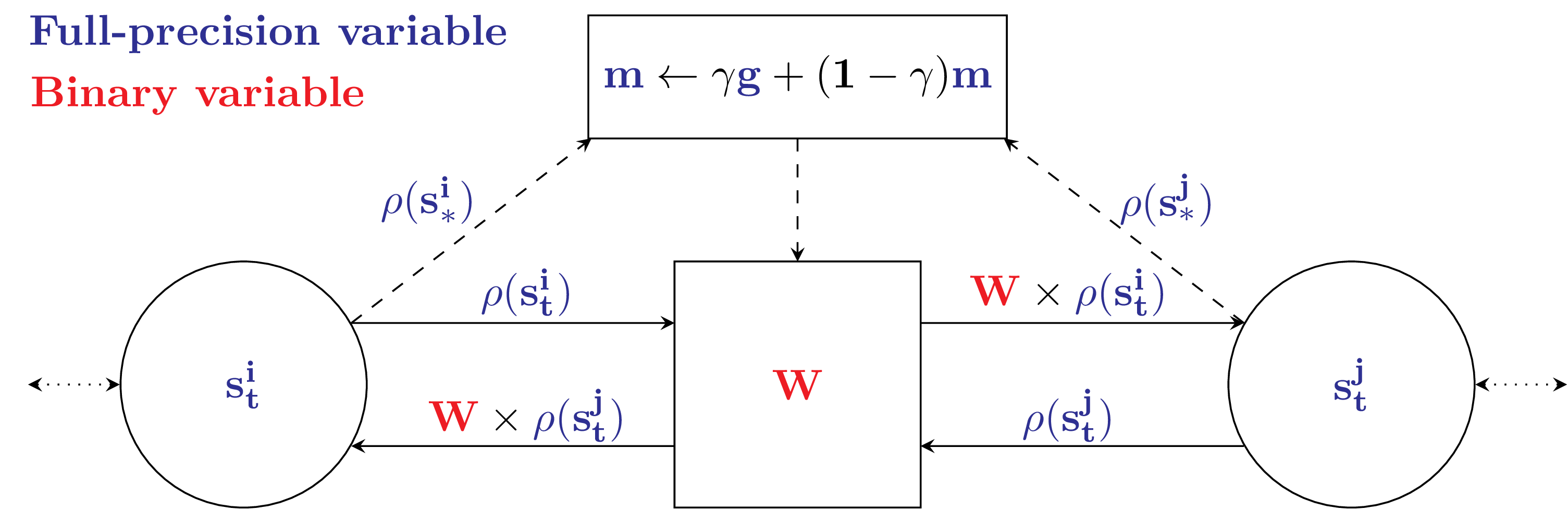}\label{fig:schema-section3}} \\
  \vspace*{-0.1in}
    \subfloat[b][Binary weights and activations (Section \ref{sec:bin-eqprop})]{\includegraphics[width=0.5\textwidth]{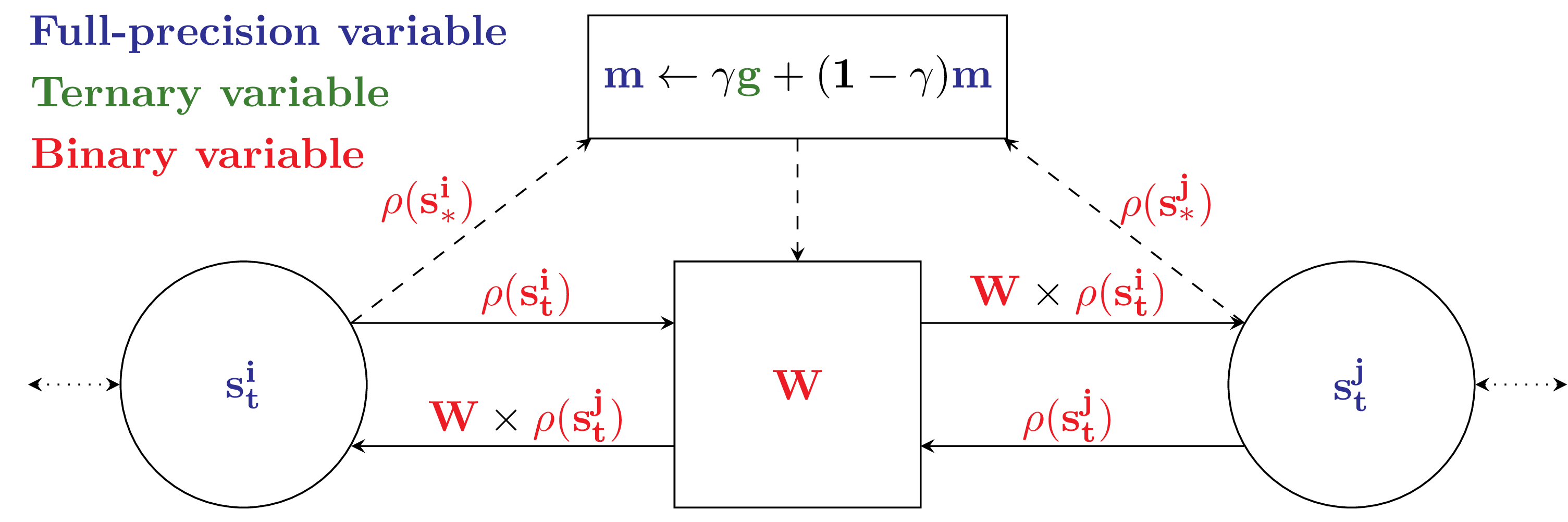}\label{fig:schema-section4}} \\
  \caption{Building blocks of binarized EP: two neurons communicate bidirectionally through a binary synapse. The color code highlights the precision of each variable: the synaptic weight (bold red) is binary and the internal state of the neurons ($s_{t}^{k}$), as well as the momentum ($m$) averaging the gradient (bold blue), are full-precision. The activations ($\rho(s_{t}^{k})$) and the equilibrium activations ($\rho(s_{*}^{k})$) are full-precision variables (blue) in Section \ref{sec:bin-W-eqprop} but binary (red) in Section \ref{sec:bin-eqprop}. The gradient estimate ($g$) prescribed by EP (bold blue) is a full-precision variable in Section \ref{sec:bin-W-eqprop} but is ternary (bold green) in Section \ref{sec:bin-eqprop}. Every neuron also has a bias which is full-precision and does not appear on the figure for clarity.} 
  \label{fig:schema-principe}
    \vspace*{-0.05in}
\end{figure}

In this paper, we address the issue of on-chip learning via EP by training dynamical systems having binary activations and weights.We first leverage the recent progress made in Binary Neural Networks (BNNs) optimization \cite{NEURIPS2019_9ca8c9b0}, to binarize the synapses in energy-based models trained by EP. The optimization of weights is performed using the inertia of the gradient. This lowers the memory required for training such systems compared to real-valued (``latent'') weights optimization, traditionally used for training BNNs. We then binarize the activation functions, yielding an easy way to compute the local gradient while supporting a digital communication between neurons. More precisely, our contributions are:
\begin{itemize}
    \item We introduce a version of EP that can learn recurrent binary weights assuming full-precision activations (Fig.~ \ref{fig:schema-section3}). For simplicity, we call this version of EP ``binarized EP''. Our implementation uses a novel weight normalization scheme directly learnable by EP. We are able to maintain an accuracy similar  to full-precision models on fully connected and convolutional architectures on MNIST. We extend these results to the CIFAR-10 task, with performance only degraded by 1.9 \%
    from that achieved with the full-precision counterpart trained by EP \cite{laborieux2020scaling}. 
    \item We extend our technique to fully binarized dynamical networks where both weights and neuron activations are binarized  (Fig. \ref{fig:schema-section4}). We demonstrate successful training on fully connected and convolutional architectures on MNIST with a slight degradation (between 0.2 and 1\%) with respect to standard EP. This ``fully binarized'' version of EP  achieves binary communication between neurons while reducing the memory required to compute the local gradient to 1 bit, making the gradient ternary. 
    \item Our code is available at: \url{https://github.com/jlaydevant/Binary-Equilibrium-Propagation}.
\end{itemize}

\vspace*{-0.05in}
\section{Background}

\paragraph{Energy-based models.}
As emphasized above, our work focuses on dynamical energy-based neural networks as opposed to purely feedforward models. More precisely, denoting $s$ the state of the neurons at a given time, $\rho(s)$ the activation function of the neurons, $x$ an input and $\theta=\{W,b\}$ the parameters of the model, we assume dynamics of the form:
\begin{equation}
    \frac{ds}{dt} = - \frac{\partial E}{\partial s}(x, s,\rho(s),\theta),
    \label{eq:def-eb}
\end{equation}
where $E(x, s,\rho(s), \theta)$ denotes an energy function describing the system of interest. Given $x$ and $\theta$, the system evolves according to Eq.~(\ref{eq:def-eb}) until reaching a steady state $s_*$ which minimizes the energy function: this constitutes the first phase. Given a target $y$ for the output layer of the system, the learning objective is to optimize the synaptic weights $\theta$ to minimize the loss:
\begin{equation}
\mathcal{L}_*=\ell(s_*, y)
\label{eq:def-loss}
\end{equation}
where $\ell$ denotes a cost function that outlines the discrepancy between $s_*$ and $y$. After learning, the system evolves to steady states of minimal prediction error. 

\paragraph{Equilibrium Propagation (EP).} While the learning objective could be optimized by backpropagating the prediction error backward in time (BPTT), EP instead proceeds with a second phase where the dynamics of Eq.~(\ref{eq:def-eb}) is changed into:
\begin{equation}
    \frac{ds}{dt} = - \frac{\partial E}{\partial s} - \beta\frac{\partial \ell}{\partial s},
    \label{eq:def-ep-second-phase}
\end{equation}
where $\beta$ denotes a scalar nudging parameter. In this way, the system evolves along Eq.~(\ref{eq:def-ep-second-phase}) towards decreasing the cost function $\ell$ until reaching a second steady state $s^\beta_*$. In their foundational paper, Scellier \& Bengio \cite{DBLP:journals/corr/ScellierB16} proved that $\mathcal{L}_*$ could be minimized using the local gradient estimate:
\begin{equation}
    \mathbf{g}_{\theta} = \frac{1}{\beta}\left(\frac{\partial E}{\partial \theta}(x, s^\beta_*, \theta) - \frac{\partial E}{\partial \theta}(x, s_*, \theta)\right),
\label{eq:ep-learning-rule1}
\end{equation}
which typically translates, for the weights of a fully connected layer, to \cite{DBLP:journals/corr/ScellierB16}:
\begin{equation}
    \Delta W_{ij} = \frac{1}{\beta}\left(\rho(s_{i, *}^\beta)\rho(s_{j, *}^\beta) - \rho(s_{i, *})\rho(s_{j, *})\right),
\label{eq:ep-learning-rule2}
\end{equation}
where $\rho$ denotes an activation function. EP has extremely attractive features for neuromorphic chip design: the same dynamics sustain both inference (Eq.~(\ref{eq:def-eb})) and error propagation (Eq.~(\ref{eq:def-ep-second-phase})), and the learning rule is local (Eq.~(\ref{eq:ep-learning-rule2})).

\paragraph{Binary Neural Networks (BNNs).} BNNs were first introduced by Courbariaux \etal \cite{courbariaux2015binaryconnect} to reduce the memory footprint and the cost of operations in feedforward neural networks at inference time, later scaled to hard visual tasks \cite{DBLP:journals/corr/RastegariORF16}. In BNNs, the weights and activations are constrained to the binary values $\{-1, + 1\}$. During BNN training, each binary weight is paired with a full-precision ``latent'' weight which undergoes weight updates. Binary weights are taken equal to the sign of the latent weights and are used for the forward and the backward passes. After training, the latent weights are discarded.

\paragraph{Binary Optimizer without latent weights (BOP).} Although latent weights in BNNs accumulate weight updates, Helwegen \etal \cite{NEURIPS2019_9ca8c9b0} suggested that they were not weights in the strictest sense (they are not used at run time) but were only meant to convey inertia for the optimization of the binary weights. Based on this insight, Helwegen \etal \cite{NEURIPS2019_9ca8c9b0} proposed a Binary Optimizer (BOP) which flips the binary weights solely based on the value of their associated momentum (without latent weights \emph{per se}): if the momentum is large enough and crosses a threshold from below, the binary weight is switched. By using one full precision variable instead of two per synapse, BOP is of definite interest to reduce the memory footprint of BNN training, which is why our work heavily relies on this technique (see Section~\ref{sec:bin-W-eqprop}). BOP has two hyperparameters:  the value of the flipping decision threshold $\tau$, and the adaptativity rate $\gamma$. The larger $\tau$, the less frequent the binary weight flips and the slower the learning. On the other hand the larger $\gamma$, the more sensitive the momentum to a new gradient signal, the more likely a binary weight flips. The BOP algorithm is summarized in Alg.~\ref{alg:bop}.

\begin{algorithm}[h]{\emph{Input}: $g$, $m$, $\textcolor{red}{\boldsymbol{W}}$, $\gamma$, $\tau$.  \\
\emph{Output}: $m$, $\textcolor{red}{\boldsymbol{W}}$.}
    \caption{BOP \cite{NEURIPS2019_9ca8c9b0}.}
    \label{alg:bop}
    \begin{algorithmic}
        \STATE $m \gets \gamma g + (1 - \gamma)m$
        \FOR{$i \in [1, d]$}
        \IF{$|m_i| > \tau$ and $\mbox{sign}(m_i) = \mbox{sign}(\textcolor{red}{\boldsymbol{W_i}})$}
        \STATE $\textcolor{red}{\boldsymbol{W_i}} \gets - \textcolor{red}{\boldsymbol{W_i}}$
        \ENDIF
        \ENDFOR
    \end{algorithmic}

\end{algorithm}
\vspace*{-0.2in}

\paragraph{Related work.} 
Spiking neural networks (SNNs) are models that compress the communication between neurons to one bit. They are thus compatible with digital and energy efficient hardware  \cite{Merolla668,spinnaker2014,trueNorth2015,loihi2018}. Most existing hardware implementations of SNNs on neuromorphic platforms use Spike Timing Dependent Plasticity (STDP) as a learning rule \cite{snnOpportunitiesPfeiffer}. Despite its low accuracy on complex tasks, the locality of STDP indeed enables compact circuits for on-chip training. This shows the importance of making EP compatible with digital hardware: its local learning rule can be implemented with compact circuits and the accuracy greatly improved compared to STDP as EP optimizes a global objective junction.

Other studies have investigated how much synapses and (or) neural activations of energy-based models could be compressed when trained by EP. Mesnard \etal \cite{mesnard2016deep}, O'Connor \etal \cite{pmlr-v89-o-connor19a} and Martin \etal \cite{martin_eqspike_2021} showed that EP can train networks where neighboring neurons communicate with spikes only. However all these techniques require full precision weights, as well as an analysis of the spike trains in order to determine the firing rates giving the gradients and are only demonstrated on MNIST or non-linear toy problems. 

Ji \& Gross \cite{ji2020towards} have studied the effect of weight and gradient quantization of an energy-based model trained by EP, showing that at least 12 or 14 bits are required to achieve less than $10\%$ test error on MNIST. Here we show that the weights (at all time) and the neural activations (at read time) can be compressed down to 1 bit only, yielding ternary gradients (at read time) and binary communication between neurons in the system. We discuss how the full-precision pre-activations and  accumulated weight momentum can be handled in a neuromorphic chip.  Finally, our work is the first to demonstrate energy-based model compression with EP on CIFAR-10. 
\section{EP Learning of Recurrent Binary Weights with Full Precision Neural Activations}
\label{sec:bin-W-eqprop}

In this section, we show that we can train dynamical systems with binary weights and full precision activations by EP with a performance on MNIST and CIFAR-10 close to the one achieved by their full-precision counterparts trained by EP \cite{NIPS2019_8930, laborieux2020scaling}. Our technique relies on the combined use of BOP described in Alg.~\ref{alg:bop} and of a proper weight normalization to avoid vanishing gradients. Therefore, we first describe how EP can be embedded into BOP (Alg.~\ref{alg:ep-bop}). Then, we propose two weight normalization schemes: one with a fixed scaling factor taken from \cite{DBLP:journals/corr/RastegariORF16}, another one with a dynamical scaling factor directly learned by EP. We show that the use of the learnt weight normalization, which naturally fits into the EP framework, considerably improves model fitting and training speed on MNIST and CIFAR-10.

\subsection{Feeding EP weight updates into BOP}
\label{ep-bop}

\paragraph{Working principle.} We explain here how to use BOP to optimize the binary synapses given the gradient computed with EP. At each training iteration, the first steps of our technique are the same as standard EP: the first phase and the second phase are performed as usual and the EP gradient estimate $g$ is obtained from the steady states $s_*$ and $s_*^\beta$ for each synaptic weight.
Thereafter, $g$ is directly fed into the BOP algorithm (Alg.~\ref{alg:bop}): for each synapse connecting neuron j to neuron i, the EP gradient estimate $g_{ij}$ conveys inertia to the synaptic momentum $m_{ij}$, and the binary synaptic weight $W_{ij}$ is flipped or not, depending on the value of $m_{ij}$. Finally, as usual in BNNs \cite{courbariaux2015binaryconnect}, the biases are full-precision and are updated with standard Stochastic Gradient Descent (SGD). We summarize all those steps in Alg. \ref{alg:ep-bop}, where we have highlighted binarized variables in bold red for clarity. With this procedure, we have a system in which the synapses are binarized at all time. In this section we use a full-precision activation function for the neurons, the hardsigmoid, and full-precision gradients. The binarization of activation functions and ternarization of gradients is addressed in Section~\ref{sec:bin-eqprop}.

\begin{algorithm}[h]{
\emph{Input}: x, y, $s$, $\beta$, $\theta=\{$\textcolor{red}{$\mathbf{W}$}, $b\}$, $\eta$, $m$, $\gamma$, $\tau$.  \\
\emph{Output}: $\theta=\{$\textcolor{red}{$\mathbf{W}$}, $b\}$, $m$.}
    \caption{EP learning of dynamical binary weights (with simplified notations). Binarized variables are in bold red. When the neural pre-activations are binarized  (Section~\ref{sec:bin-eqprop}), the EP gradient estimate $g$ is ternarized (in bold green), otherwise full precision (Section~\ref{sec:bin-W-eqprop}).}
    \label{alg:ep-bop}
    \begin{algorithmic}
        \STATE Free phase:
        \FOR{$t \in [1, T]$}
        \STATE $s \gets s - dt\times \frac{\partial E(\text{x}, s, \mathbf{\textcolor{red}{W}}, b)}{\partial s}$
        \ENDFOR
        \STATE $s_* \gets s$
        \STATE Nudged phase:
        \FOR{$t \in [1, K]$}
        \STATE $s \gets s - dt\times \frac{\partial E(\text{x}, s, \mathbf{\textcolor{red}{W}}, b)}{\partial s} - \beta \times \frac{\partial \ell (\text{y},y)}{\partial s} $
        \ENDFOR
        \STATE $s^{\beta}_* \gets s$
        \STATE Compute EP gradient with $s_*^\beta$ and $s_*$:
        \STATE \quad ${\color{OliveGreen}\mathbf{g}} \gets -\frac{1}{\beta}\left(\frac{\partial E}{\partial \theta}(\text{x}, s^\beta_*, \mathbf{\textcolor{red}{W}}, b) - \frac{\partial E}{\partial \theta}(\text{x}, s_*, \mathbf{\textcolor{red}{W}}, b)\right)$
        \STATE Apply BOP (Alg.~(\ref{alg:bop})):
        \STATE \quad $m$, \textcolor{red}{$\mathbf{W}$} = ${\rm BOP}(g, m, \mathbf{\textcolor{red}{W}}, \gamma, \tau)$
        \STATE Update biases with SGD:
        \STATE \quad $b \gets b + \eta \times g$
    \end{algorithmic}

\end{algorithm}
\vspace*{-0.2in}

\paragraph{Hyperparameter tuning.} Similarly to \cite{NEURIPS2019_9ca8c9b0}, we monitor the number of weight flips per epoch and layer-wise in order to tune the hyperparameters of BOP, using the metric:
\begin{equation}
    \pi_{\rm epoch}^{layer}=\log\left(\frac{\text{Number of flipped weights}}{\text{Total number of weights}} + e^{-9}\right)
    \label{eq:metric-bop}
\end{equation}
Heuristically, $\pi_{\rm epoch}^{layer}$ reflects a trade-off between learning speed (high $\pi_{\rm epoch}^{layer}$) and stability (low $\pi_{\rm epoch}^{layer}$). We measure $\pi_{\rm epoch}^{layer}$ in the regions of $\gamma$ and $\tau$ where learning performs well, and use this value of $\pi_{\rm epoch}^{layer}$ in return as a criterion to tune $\gamma$ and $\tau$ on new models.

\subsection{Normalizing the Binary Weights with a fixed scaling factor}
\label{binarize-W-fixed}

When binarizing synaptic weights to $\pm 1$, neural activities may easily saturate to regions of flat activation function, resulting in vanishing gradients. It is especially true with the hardsigmoid activation function often used with EP. 
Batch-Normalization \cite{DBLP:journals/corr/IoffeS15} used by Courbariaux \etal \cite{courbariaux2015binaryconnect} and Hubara \etal \cite{NIPS2016_6573} helps with this issue by recentering and renormalizing activations by computing the batch statistics. Batch-Normalization has been introduced in recurrent neural networks such as LSTMs to process sequence tasks \cite{laurent2016batch} but it does not translate directly to energy-based models. The normalization scheme should indeed itself derive from an energy function in order to be learnable, which restricts the choice of candidate normalizations. However, the goal in convergent dynamical systems processing static inputs is not to center neural activations at every time step, but rather at their steady state. Moreover, using batch-based weight normalization schemes is far from straightforward from a hardware prospective. For this purpose, we first normalize the binary weights with a static scaling factor. 

\paragraph{Static XNOR-Net weight scaling factor.} In the design of their XNOR-Net model, Rastegari \etal \cite{DBLP:journals/corr/RastegariORF16} introduced a scaling factor to minimize the difference between the binary synapses and the corresponding set of full-precision ``latent'' weights at each layer. This scaling factor is updated at each training iteration and depends on the size of two adjacent layers and on the magnitude of these latent weights. The scaling factor reads in our context:
\begin{equation}
    \alpha_{n, n + 1}=\frac{||w_{n,n+1}^{\rm init}||_{1}}{{\rm dim}(w_{n, n +1}^{\rm init})}
    \label{eq:init-alpha}
\end{equation}
where $n$ is the index of a layer, $w_{n,n+1}^{\rm init}$ are the full-precision random weights used to initialize the binary weights. Using this scaling factor, we initialize each binary weights, layer by layer, as $W_{n, n + 1} = \pm \alpha_{n, n + 1}$.
Contrarily to XNOR-Net where the scaling factor is updated at each forward pass, we first keep the scaling factor fixed to its initial value throughout training. We show in Section \ref{annexe:scaling-factor} that this 
scaling factor is crucial to train recurrent binary weights by EP.

\begin{table*}[ht!]
 \caption{Error of EP and BPTT on networks having binary synapses with a fixed or a dynamical scaling factor - Results are reported as the mean over 5 trials $\pm$ 1 standard deviation - Benchmark performances are taken from \cite{NIPS2019_8930, laborieux2020scaling}}
  \centering
  \begin{tabular}{lcccccccccc}
    \toprule
    & & & \multicolumn{4}{c}{\textbf{EP - Binary Synapses}} & & \textbf{EP}  & & \textbf{BPTT}\\
     \cline{4-7}
     \cline{9-9}
     \cline{11-11}
      & & & \multicolumn{2}{c}{Fixed $\alpha$} & \multicolumn{2}{c}{Dynamical $\alpha$} & & Benchmark  & &  Binary Synapses\\
     \hline
     Dataset & Model & & Test & Train & Test & Train & & Test & & Test\\
    \midrule
    MNIST & (1fc) & & 2.07 (0.02) & $0.77$ & {\bfseries 1.7 (0.04)} &  {\bfseries 0} & & 2.00 & & 2.14 (0.06) \\
    MNIST & (2fc) & & 2.48 (0.08) & $0.29$ & {\bfseries 2.28 (0.13)} & {\bfseries 0}&  & 1.95 & & 2.38 (0.07) \\ 
    MNIST & (conv) & & $0.85 (0.11)$ &$0.46$  & $\mathbf{0.88 (0.06)}$ & $\mathbf{0.05}$ & & 1.05 & & 0.97 (0.03) \\
    CIFAR-10 & (conv) & & $16.8 (0.3)$ &  6.9 & $\mathbf{15.66 (0.28)}$ & $\mathbf{5.54}$ & & 13.78 & & 14.45 (0.12)  \\
    \bottomrule
  \end{tabular}
  \label{tab:results-table-bin-W-EP}
  \vspace*{-0.1in}
\end{table*}

\paragraph{Results.} We investigate fully connected architectures 
(with one and two hidden layers) 
on MNIST and convolutional architectures (with two and four convolutional layers)
on the MNIST and CIFAR-10 datasets. We employ prototypical models to speed up training as in \cite{NIPS2019_8930}. Our results (Table \ref{tab:results-table-bin-W-EP} - ``EP - Binary Synapses'') are benchmarked against those of full precision models (Table \ref{tab:results-table-bin-W-EP} - ``EP - Benchmark'') and those obtained by BPTT+BOP (Table \ref{tab:results-table-bin-W-EP}  - ``BPTT - Binary Synapses''). Note that for a given architecture, the number of neurons we used per layer may not be the same as in reference architectures -- see \ref{binarize-W-hardware} and  Appendix~\ref{annexe:simulations-details} for details.

Overall, Table \ref{tab:results-table-bin-W-EP} shows that the normalization of weights with a fixed scaling factor allows EP with binary synapses to perform comparably to full-precision models trained across different fully connected and convolutional architectures, on MNIST and CIFAR-10. The fully connected architecture which has one hidden layer trained on MNIST shows no statistically significant loss of performance compared to full-precision counterpart trained by EP. This architecture with binary synapses reaches the same accuracy if trained by (EP+BOP) or by (BPTT+BOP). The fully connected architecture having two hidden layers trained on MNIST with fixed scaling factors shows $~0.5\%$ performance degradation compared to full-precision models trained by EP. For the convolutional architecture trained on MNIST, we can even observe a slightly better training and testing (-0.2\%) errors on model with binary synapses compared to full precision models trained by EP. We explain this improvement by the cumulative use of the randomization of $\beta$ and of the regularization effect induced by the binarized architecture itself \cite{courbariaux2015binaryconnect}. 
Furthemore, the training framework (EP+BOP) achieves a similar accuracy as the framework (BPTT+BOP). Finally, the performance of our convolutional model trained on CIFAR-10 is $\sim 3\%$ less than the one of Laborieux \etal \cite{laborieux2020scaling}, using the same architecture. Also, the network trained by (EP+BOP) shows only 2.5\% degradation of the accuracy compared to the same network trained by (BPTT+BOP).

\subsection{Normalizing the Binary Weights with a learnt scaling factor}
\label{binarize-W-learnt}

\begin{figure}[ht!]
  \centering
  \hspace*{-0.1in}
  \subfloat[a][MNIST]{\includegraphics[width=0.5\textwidth]{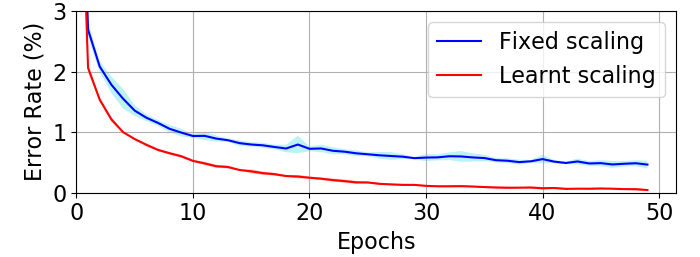}\label{fig:sub1}} \\
  \vspace*{-0.15in}
  \hspace*{-0.1in}
  \subfloat[b][CIFAR-10]{\includegraphics[width=0.5\textwidth]{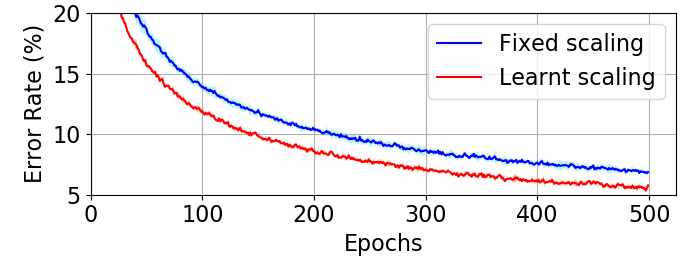}
  \label{fig:sub2}}
  \vspace*{-0.1in}
  \caption{Average training error as a function of the number of epochs for a convolutional architecture with binary synapses trained on MNIST with a static (blue curve) or a dynamical (red curve) scaling factor. Curves averaged over 5 trials $\pm 1$ standard deviation.} 
  \label{fig:fixed-learnt-alpha}
  \vspace*{-0.2in}
\end{figure}

\paragraph{Dynamical weight scaling factor learned by EP.}
Using fixed scaling factors gives high, yet sub-optimal accuracies (see Fig. \ref{fig:test_error_vs_alpha} in the Appendix). Bulat \& Tzimiropoulos
\cite{DBLP:journals/corr/abs-1909-13863} show that the scaling factor can be learnt by backpropagation to extend XNOR-Nets. Here we derive a learning rule for the scaling factor with the help of the theorem of Scellier \& Bengio \cite{DBLP:journals/corr/ScellierB16} to ensure that it provides a gradient estimate of the loss $\mathcal{L}_*$ defined in Eq.~(\ref{eq:def-loss}). 
The reasoning to derive this learning rule is the following. We split the binary weights in two parts: $W_{n, n+1} \gets \alpha_{n,n+1} \times w_{n, n+1}^{bin}$ where $w_{n, n+1}^{bin}$ are the binary weights scaled to $\pm1$ and the scaling factors $\alpha_{n,n+1}$ are initialized as when they are fixed. The resulting dynamics still derives from an energy function, and one can derive  a learning rule for $\alpha_{n, n + 1}$ which reads as:
\begin{equation}
\Delta \alpha_{n, n + 1}(\beta) =-\frac{1}{\beta}\left(\frac{\partial E}{\partial \alpha_{n, n + 1}}(s_*^\beta) - \frac{\partial E}{\partial \alpha_{n, n + 1}}(s_*)\right),
\end{equation}
so that $\alpha_{n, n+1}$ is learned like any other network parameter, and $\lim_{\beta \to 0}\Delta \alpha_{n, n + 1}(\beta) = - \frac{\partial \mathcal{L}_*}{\partial \alpha_{n, n + 1}}$. In Section \ref{annexe:scaling-factor} the learning rules for fully connected and convolutional architectures are derived in all the settings of EP.

\paragraph{Results.} Fig. \ref{fig:fixed-learnt-alpha} illustrates on a convolutional architecture the gain in performance obtained by learning the scaling factor. Globally, this technique systematically results in faster learning and better model fitting across all the models, and almost always in better generalization as observed in Table~\ref{tab:results-table-bin-W-EP}. Learning the scaling factor by EP is thus a powerful alternative to Batch-normalization in convergent dynamical systems as highlighted by Fig. \ref{fig:fixed-learnt-alpha}. 

\subsection{Hardware implementation}
\label{binarize-W-hardware}

\paragraph{Memory gain at run time.} As often observed in binarized architectures \cite{courbariaux2015binaryconnect, NIPS2016_6573}, we achieve accuracy similar to the one of full-precision models at the price of having 8 times more hidden neurons in fully connected architectures (see Fig. \ref{fig:error_vs_hiddenNeurons} of Appendix~\ref{subannexe: bin-synapses} for a more detailed analysis). In convolutional architectures, we have used at most the same number of output feature maps than their full precision counterparts for computational efficiency. After training, our models use 2 and 7.5 less memory for the synapses for fully connected architectures on MNIST (for two and one hidden layers respectively), 9 and 54 less for the convolutional architectures used on MNIST and CIFAR-10 respectively. In hardware, these binary weights can be stored in digital memories \cite{thakur_large-scale_2018} or using nanoscale memristors \cite{hirtzlin2020}. 

\paragraph{Memory requirements at train time.} The memory requirements for training should be subject to a more careful treatment. Inertia-based optimization (BOP) requires a single full-precision variable: the momentum, compared to the latent-weight counterpart often trained by elaborate optimization techniques such as (SGD + Momentum) which uses at least 2 full-precision variables: the latent weight and the momentum. Inertia-based optimization thus reduces by at least a factor 2 the required memory for training. Furthermore, the memory required for storing the momentum of BOP could be implemented by non-standard memories. In fact, the discrete time update of the momentum (Alg. \ref{alg:bop}) can be rewritten into a continuous time update rule which reads: 
\begin{equation}
\frac{dm}{dt}+\gamma\cdot m(t)=\gamma\cdot g(t)
\end{equation} which naturally appears as the differential equation describing the evolution of the voltage of a capacitor. Capacitors are CMOS-compatible, and highly linear which makes them well suited for storing full-precision variables \cite{ambrogio2018}. They can thus be used to store the inertia, thereby lowering the memory requirement to one capacitor per binary weight and more globally lowering the memory required for training.

\section{EP Learning of Recurrent Binary Weights with Binary Neural Activations}
\label{sec:bin-eqprop}

The techniques presented in the previous section use full-precision neural activations. However, it is highly preferable to rely on binary activation values in hardware. Binary read and write errors can indeed be accommodated without too much circuit overhead in neuromorphic systems \cite{hirtzlin2019hybrid} and binary values are easier to pass between spatially distant hardware neurons \cite{thakur_large-scale_2018}.
In this section, we show that we can train dynamical system with binary weights and binary activations by EP, resulting in a performance on MNIST again approaching full-precision models on fully connected and convolutional architectures. Our implementation relies on two main components: the choice of a proper activation function to binarize neural pre-activations, and output layer augmentation. Combining these two techniques, we can design dynamical systems which are sensitive to error signals despite threshold effects and can compute ternary gradients in return. The corresponding pseudo-algorithm is the same as Alg.~\ref{alg:ep-bop}, except that the gradient estimate $g$ is now ternarized.

\subsection{Convergent neural networks with binary activations.} 
\paragraph{Ternarizing EP gradients.} We can note from Eq.~(\ref{eq:ep-learning-rule2}) that the precision of the gradient $g$ estimate provided by EP is typically determined by the choice of the activation function $\rho$. For instance, if $\rho$ outputs binary values $\{0, 1\}$, we immediately see from Eq.~(\ref{eq:ep-learning-rule2}) that, for the parameters $\theta=\{W,b\}$, the gradients has values:
\begin{equation}
\mathbf{g}_\theta\in \left\{-\frac{2}{\beta}, 0, \frac{2}{\beta}\right\}.
\label{eq:dW-tern}
\end{equation}
In practice $\beta = 2$ works well, resulting in $\times 40$ gradient compression compared to $64$-floating point resolution.

However, binarizing neural activations comes at several costs for the dynamics of the neurons. The energy function of the system subsequently outputs a semi-discrete variable which affects the dynamics of each neuron non trivially: if the updates of neuron activations are simultaneous, the dynamics of the system may not converge \cite{Cheung:87}. In particular, this precludes the use of prototypical models \cite{NIPS2019_8930}, that can be employed to speed up training as we did in the previous section. Therefore, we must use standard energy-based models as in \cite{DBLP:journals/corr/ScellierB16} so that binary activations are updated only when the full-precision pre-activations reach the threshold of the activation function, thus non simultaneously. We next detail some empirical properties the binary activation of the neurons should have in order to define convergent dynamics.

\paragraph{Binarizing neural activations into $\{0, 1\}$.}

While we binarize weights to opposite signs, we found that using the sign activation function to binarize neural activations into $\{-1, 1\}$, as usually done with BNNs to implement MAC operations with XNOR gates, entails non-convergent dynamics. This confirms previous findings on EP which emphasized the importance of bounding the neural activations between 0 and 1 to help dynamics convergence using the hardsigmoid activation function $\rho(s) = {\rm max}(0, {\rm min}(s, 1))$
\cite{DBLP:journals/corr/ScellierB16}. 
Therefore, our proposal here is to use the Heavyside step function shifted by 0.5:
\begin{equation}
    \rho(s) = {\rm H}\left(s - \frac{1}{2}\right)
    \label{eq:heavi}
\end{equation} where $H(x) = 1$ if $x \geq 0$, 0 otherwise. However, the energy based dynamics of our models requires to gate $\frac{\partial E}{\partial \rho}$  by the derivative of $\rho$ as Eq. \ref{eq:def-eb} rewrites as:
\begin{equation}
    \frac{ds}{dt} = -\frac{\partial E}{\partial s}(x,s,\theta)-\frac{\partial \rho(s)}{\partial s}\frac{\partial E}{\partial \rho(s)}(x,\rho(s),\theta).
    \label{eq:def-eb-2}
\end{equation}
Noting that Eq. \ref{eq:heavi} is obtained by asymptotically sharpening the narrowed hardsigmoid around $\frac{1}{2}$ within $[\frac{1}{2} - \sigma, \frac{1}{2} + \sigma]$ denoted $\hat{\rho}$, namely:
\begin{equation}
    \rho(s) = \lim_{\sigma \to 0} \hat{\rho}(s,\sigma) = {\rm H}\left(s - \frac{1}{2}\right)
    \label{eq:heavi-lim}
\end{equation}
we propose to substitute
the derivative of $\rho$ as:
\begin{equation}
    \frac{\partial \hat{\rho}(s, \sigma)}{\partial s} \approx \frac{\partial \boldsymbol{\rho}(s)}{\partial s}=
    \begin{cases}
      \frac{1}{2\sigma} & \text{if } \left|s - \frac{1}{2}\right| \leq \sigma \\
      0 & \text{else}
    \end{cases} 
    \label{eq:def-pseudo-derivative}
\end{equation} where $\sigma$ is a parameter discussed in Appendix \ref{annexe:simulations-details}.
Therefore, denoting $\hat{\rho}' = \partial_s \hat{\rho}$ for simplicity, the free dynamics of $s$ can be approximated as:
\begin{equation}
    \frac{ds}{dt} \approx - \frac{\partial E}{\partial s} - \hat{\rho}'(s)\frac{\partial E}{\partial \rho}
\label{eq:dynamics-bin-EP2}    
\end{equation}

\subsection{Augmenting the Error Signal to Nudge Neurons with Binary Activations}
\label{error-signal}

\paragraph{Binarization of activations can prevent the propagation of errors.}
As the system sits at rest at the end of the first phase of EP, upon nudging the output layer by the prediction error, the motion of the system during the second phase of EP encodes error signals \cite{scellier2019equivalence, NIPS2019_8930}. Therefore during the second phase, a given neuron $i$ needs to have its activation function change from $\rho(s_{*, i})$ to a distinct $\rho(s_{*, i}^\beta)$ to compute the error gradient locally and transmit it backward to upstream layers. However, when using a discontinuous activation function like defined in Eq.~(\ref{eq:heavi-lim}), we may have $\rho(s_{*, i}) = \rho(s_{*, i}^\beta)$ if the pre-activation $s_i$ of the neuron  moves less than the value of the activation threshold of $\rho$, thus zeroing the error signal, or equivalently vanishing gradients. Consequently, we need to ensure that for a sufficient number of neurons $i$: 
\begin{equation}
    \Delta s_i = |s^{\beta}_{*, i}-s_{*, i}|>\frac{1}{2}
    \label{eq:condition-bin-neurons}
\end{equation}

In order to satisfy Eq.~(\ref{eq:condition-bin-neurons}) for a sufficient number of neurons, we propose to increase the error signal by augmenting the output layer so that each prediction neuron is replaced by $N_{\rm per class}$ neurons per class, inflating the output layer from $N_{\rm classes}$ to $N_{\rm classes} \times N_{\rm per class}$. We choose $N_{\rm per class}$ in such a way that the number of output neurons matches approximately the number of neurons in the penultimate hidden layer: $N_{\rm per classes} \approx \frac{N_{\rm penultimate}}{N_{\rm classes}}$.  
In this way, the output layer delivers a large and redundant initial error signal that can push neurons beyond the activation threshold of $\rho$ and propagate across the whole architecture.
Our solution is reminiscent of the the use of auxiliary output neurons in \cite{bartunov2018assessing}, albeit with a very different motivation.

\subsection{Results}

We investigate here fully connected (1 and 2 hidden layers) and convolutional architectures on MNIST. The first layer receives full-precision inputs from the input layer and binary inputs from the next layer. For a given architecture, the number of neurons used per layer is different for both situations: for the fully connected architectures we use 8192 neurons per hidden layer and the two convolutional layers of the convolutional architecture have respectively 256 and 512 channels - see Appendix \ref{annexe:simulations-details} for more details. We use a randomized sign for $\beta$ as prescribed by Laborieux \etal \cite{laborieux2020scaling} to improve the gradient estimate given by EP for all simulations except for the fully connected architecture with two hidden layers where we only use $\beta>0$.

\begin{figure}[!ht]
\begin{center}
\hspace*{-0.1in}
   \includegraphics[width=0.5\textwidth]{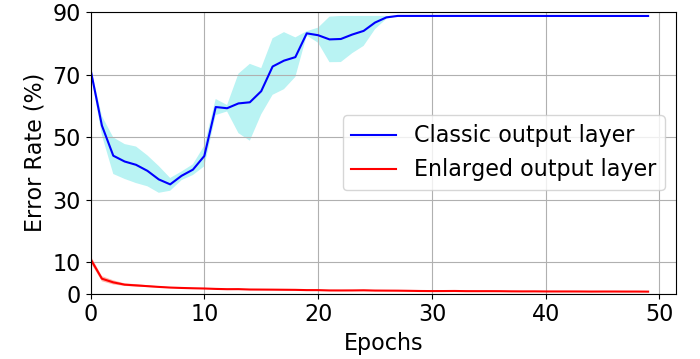}
\end{center}
\vspace*{-0.2in}
  \caption{Average training error as a function of the number of epochs for a convolutional architecture with binary synapses and binary activations trained on MNIST with a classic output layer (10 output neurons - blue curve) or an enlarged output layer (700 output neurons - red curve). 
  Blue curves are averaged over 2 trials $\pm 1$ standard deviation - Red curves are averaged over 5 trials $\pm 1$ standard deviation.}
  \label{fig:large-output-vs-small}
\end{figure}

Fig \ref{fig:large-output-vs-small} shows for the convolutional architecture a trend observed for all models:  when using 10 neurons in the output layer, training fails (blue curve) while it succeeds upon augmenting the output layer. It is here augmented by a factor 70 (red curve) which is required for the number of neurons in the output layer to match the number of input neurons that the last convolutional layer receives from the penultimate convolutional layer: we multiply the number of channels in the penultimate convolutional layer (256) by the kernel size (5×5) and divide the result by the max pooling kernel size (3×3) which gives $\sim$ 700 output neurons).

\label{results-bin}
\begin{table}[ht!]
 \caption{Error achieved by EP with binary synapses \& activations, and fixed scaling factors $\alpha$ - Results are reported as the mean over 5 trials $\pm$ 1 standard deviation.}
  \centering
  \begin{tabular}{llcc}
    \toprule
     & & \multicolumn{2}{c}{\textbf{Fully binarized EP} }
     \\
    \hline
    Dataset & Model & Test & Train \\
    \midrule
    MNIST & (1fc) & 2.83 (0.06) & 0.2 \\
    MNIST  & (2fc) & $3.03 (0.03)$ & $ 0.84 (0.17)$ \\
    MNIST  & conv & $1.14 (0.08)$ & $0.67 (0.04)$ \\ 

    \bottomrule
  \end{tabular}
  \label{tab:results-table-bin-EP}
  \vspace*{-0.2in}
\end{table}

\paragraph{Performance.} The results obtained on MNIST with fixed scaling factors are summarized in Table~\ref{tab:results-table-bin-EP}. On the fully connected architectures, the accuracy approaches those obtained with binary synapses and full-precision activations, with a slight degradation of 0.8\% for one hidden layer and 0.6\% for two hidden layers (see Table~\ref{tab:results-table-bin-W-EP}). The degradation is slightly enhanced when we compare with the full-precision counterpart trained by EP where the performance is degraded by 0.8\% for one hidden layer but 1\% for two hidden layers. We account the degraded performance of the architecture which has two hidden layers by the fact that we use $\beta >0$ which makes the estimation of the gradient less accurate than when estimated with the sign of $\beta$ random. We used $\beta >0$ because we found that reaching the second equilibrium point with $\beta<0$ is possible but very long to get in practice with a classic nudge. For the convolutional architecture trained on MNIST, we also report a performance only 0.2\% below the system which has binary synapses and full-precision activations but within the error bars of the one achieved by full-precision models as reported by \cite{NIPS2019_8930}. We think that optimizing the nudging strategy could improve the error obtained with two or more hidden layers and will be key in the future for scaling to CIFAR-10.

\subsection{Gains for hardware}
When binarizing the activation in addition to the synapses, we had to increase the number of neurons in each layer compared to full-precision models: by 16 for the fully connected layers resulting in 8192 neurons per hidden layer and by 8 for the convolutional architecture which has 256 and 512 channels per respective layer, to get accuracy approaching reported results with full-precision architectures. But considerable gains in terms of memory and computing are achieved due to the way the gradient is computed. 

The gradient estimate $\mathbf{g_{ij}}$ is indeed now ternary (Eq. \ref{eq:dW-tern}), and can be easily computed with the subtraction of 2 AND operations. With the notations of Eq. \ref{eq:ep-learning-rule2} it decomposes as: one AND operation between $s_{i,*}$ and $s_{j,*}$ and another one between $s_{i,*}^{\beta}$ and $s_{j,*}^{\beta}$, which amounts to only 5 elementary operations including the subtraction. In terms of memory, neurons only have to store 1 bit as the first equilibrium state. That way, the communication between neurons is not only binarized, but in addition, compared to previous works on EP achieving binary communication through spikes \cite{mesnard2016deep,pmlr-v89-o-connor19a,martin_eqspike_2021}, our method drastically reduces the memory to compute the gradient. Indeed, spikes need to be stored for several time steps to get an estimation of the firing rate of each neuron, resulting in heavy memory requirements.

The computation of the MAC operation is also simple. It cannot be obtained as in standard BNNs with a single XNOR gate and popcount because this operation does not match our choice of binary activations (0/1) and binary weights (-1/1). However, it only requires the subtraction of the popcount of 2 AND gates, using simpler logical gates, and only doubling their total number compared to usual BNNs. 

Despite the fact that we need to enlarge the output layer depending on the architecture, we show in Appendix \ref{annexe:error-signal-prediction} that probing the state of one single neuron per class in the output layer is sufficient to obtain almost the same accuracy than when measuring the states of all the output neurons, which is beneficial for lowering the energy consumption of hardware (Figs. \ref{fig:MNIST-1fc-8192-100-diff}-\ref{fig:MNIST-2fc-8192-8000-diff}). 

In addition, contrarily to BP performed in conventional BNNs where the input of each layer is stored between the forward and the backward passes in order to compute the full-precision gradient, here we only need to store the 1 bit activation after each phase in order to compute the gradient, which drastically reduces the memory requirements of the model for training by a factor 40. The current implementation of binary EP on GPUs, however, still relies on full-precision neuron state $s(t)$ variables. In future implementation of binarized EP on dedicated hardware, these neuron states can be implicitly encoded through the dynamics of nano-devices, thus solving this issue \cite{ambrogio2018}. 

\section{Conclusion}
\label{conclusion}
\vspace*{-0.05in}
As a conclusion, we provide here a binarized version of EP that exhibits only a slight degradation of accuracy compared to full-precision models. 
This version of EP offers the possibility of training on-chip BNNs with compact circuitry because the hardware required for training is the same as for inference, whereas current BNNs are trained on conventional hardware, before being transferred to compact, low-energy chips. Finally, the version of EP with binary synapses and full precision activations is of major interest for future fast, low power hardware built on emerging devices. Joint development of EP and hardware will be critical for adapting EP to larger data sets.

\vspace*{-0.05in}
\section{Acknowledgments}
\vspace*{-0.05in}
This work was supported by the European Research Council ERC under Grant No. bioSPINspired 682955 and Grant No. NANOINFER 715872.

{\small
\bibliographystyle{ieee_fullname}
\bibliography{main}
}

\clearpage
\onecolumn
\appendix

\section{Organization of the Supplementary Materials}
The Supplementary Materials are organized as follows. We first derive the dynamics and learning rules for the weights and biases in the energy-based and prototypical settings of EP in Section \ref{annexe:ep-fc-settings} and  Section \ref{annexe:ep-conv-settings} respectively. Then, we give more details about the scaling factor: chosen fixed or dynamical, in Section \ref{annexe:scaling-factor}. We discuss why the error signal has difficulties to flow in the system when the neurons have a binary activation function in Section \ref{annexe:error-signal}. We finally detail all the software implementations in Section \ref{annexe:simulations-details}. 

More precisely one can find:

\begin{itemize}
    \item Experimental evidence showing the importance of the scaling factor (fixed in this section) for training systems with binary weights (Section \ref{annexe:scaling-factor}).
    \item The derivation of the learning rule for the scaling factor, together with a description of the results obtained with this training procedure, showing the acceleration that the learnt dynamical scaling factor provides  (Section \ref{annexe:scaling-factor}).
    \item An empirical demonstration that  the prediction is efficiently computed with an enlarged output layer for binary neural networks (Section \ref{annexe:error-signal}).
    \item Training curves and their flipping metric (defined in Eq. (\ref{eq:metric-bop})) monitored over epochs (Section \ref{annexe:simulations-details}).
\end{itemize}

\section{Training Fully Connected Layers Networks with Equilibrium Propagation}
\label{annexe:ep-fc-settings}
In this section, we describe and define all the operations we used to train with EP a fully connected neural network 
recurrently connected through bidirectional synapses. We describe the dynamics and the underlying learning rules for the weights and the biases in the energy-based and prototypical settings. The units of the system are denoted $\mathbf{s}=\{h,y\}$ where $h$ are the hidden units and $y$ are the output units. The variable y is the one-hot encoded target vector. The inputs x are always clamped and are static.

\subsection{Energy-Based Settings}
In the energy based settings, we introduce an energy function for the network, that defines the neuron dynamics during the two phases of EP. We then derive the learning rules from the energy function.

\subsubsection{Energy Function}
We consider the following energy function \cite{DBLP:journals/corr/ScellierB16}:
\begin{equation}
    E(s, \rho(s), \theta=\{W,b\}):=\frac{1}{2}\sum_{i}s_{i}-\frac{1}{2}\sum_{i\ne j}W_{ij}\rho(s_{i})\rho(s_{i})-\sum_{i}b_{i}\rho(s_{i})
    \label{eq:energy-function-annexe}
\end{equation}
where $\rho$ is the activation function of the neurons, $W_{ij}$ the weight connecting the unit $s_{i}$ to $s_{j}$ and reciprocally as synapses are symmetric for the system to converge and $b_{i}$ the bias of unit $s_{i}$.

We also  define $\ell$ the cost function describing how far are the output units of the system ($\hat{y}$) from their target state (y). We usually employ the mean squared error as a cost function with EP:

\begin{equation}
    \ell(s, \text{y})=MSE(s,\text{y}):=\frac{1}{2}\sum||\text{y}-\hat{y}||^{2}
    \label{eq:mse-cost-function}
\end{equation}

where $\text{y}$ denotes a given target output. 

\subsubsection{Dynamics}
The dynamics of neurons in the free phase evolve according to the energy function $E$:
\begin{equation}
    \frac{ds}{dt}=-\frac{\partial E}{\partial s}
    \label{eq:ep-dynamcis-annexe}
\end{equation}

which  translates for the neuron $i$ and the energy function $E$ defined in Eq.\ref{eq:energy-function-annexe} as:

\begin{equation}
    \frac{ds_{i}}{dt}=-s_{i}+\rho'(s_{i})(\sum_{j\neq i}W_{ij}\rho(s_{j})+b)
    \label{eq:dynamics-free-phase}
\end{equation}

The system eventually settles to a fixed steady state $s_{*}$.

During the nudged phase the dynamics differs from the free phase as the neurons now evolve to decrease the cost function $\ell$:

\begin{equation}
    \frac{ds}{dt}=-\frac{\partial E}{\partial s}-\beta\frac{\partial \ell}{\partial s}
    \label{eq:ep-dynamcis-nudged-annexe}
\end{equation}

which translates for the hidden unit $h_{i}$, the output unit $\hat{y}_{i}$ and the target y to: 

\begin{equation}
    \left\{
    \begin{array}{l}
    \frac{dh_{i}}{dt}=-h_{i}+\rho'(h_{i})(\sum\limits_{j\neq i}W_{ij}\rho(s_{j})+b_{i})\\
    \frac{d\hat{y}_{i}}{dt}=-\hat{y}_{i}+\rho'(y_{i})(\sum\limits_{j\neq i}W_{ij}\rho(h_{j})+b_{i}) + \beta\times(\text{y\textsubscript{i}}-\hat{y}_{i})
    \end{array}
    \right.
    \label{eq:dynamics-nudged-phase}
\end{equation}

The systems eventually reaches a second steady state denoted $s_{*}^{\beta}$.

\subsubsection{Learning Rule}
Scellier \& Bengio
\cite{DBLP:journals/corr/ScellierB16} showed that the gradient of the loss $\mathcal{L}_*$ (defined in Eq.~(...)) with respect to any parameter in the system can be approximated by the derivative of the energy function $E$ with regard to the parameter evaluated at the two equilibrium points $s_{*}$ and $s_{*}^{\beta}$: 

\begin{equation}
-\frac{\partial \mathcal{L}_*}{\partial \theta} = \lim_{\beta \to 0} \frac{1}{\beta}\left(\frac{\partial E}{\partial \theta}(x, s^\beta_*, \theta) - \frac{\partial E}{\partial \theta}(x, s_*, \theta)\right)
\label{eq:annexe-ep-learning-rule}
\end{equation}

In the energy-based settings, the resulting learning rules for the weights and biases are expressed as a function of the two steady states:
\begin{equation}
\left\{
    \begin{array}{l}
    \Delta W_{ij}=\frac{1}{\beta}(\rho(s_{i,*}^{\beta})\rho(s_{j,*}^{\beta})-\rho(s_{i,*})\rho(s_{j,*}))\\
    \Delta b_{i}=\frac{1}{\beta}(\rho(s_{i,*}^{\beta})-\rho(s_{i,*}))
        \end{array}
    \right.
    \label{eq-learning-rule-EB-EP}
\end{equation}

\subsection{Prototypical Settings}
Ernoult \etal \cite{NIPS2019_8930} introduced the prototypical settings for EP where the dynamics no longer derived from an energy function in a continuous-time setting but more generally from a scalar primitive in a discrete-time setting. As in Ernoult \etal \cite{NIPS2019_8930}, we chose a dynamics close to the one of conventional RNNs. We then write a primitive function from which the dynamics derives. Finally we obtain the learning rules from the primitive function.

\subsubsection{Dynamics}
We choose the same discrete time dynamics as in \cite{NIPS2019_8930}:
\begin{equation}
\left\{
    \begin{array}{l}
    h_{i}^{t+1} =  \rho(\sum\limits_{j}W_{ij}s_{j}^{t}+b) \\
    y_{i}^{t+1} =  \rho(\sum\limits_{j}W_{ij}h_{j}^{t}+b) + \beta\times(\text{y\textsubscript{i}}-\hat{y}_{i}) \text{ where $\beta=0$ during the free phase}
    \end{array}
\right.
\label{eq:prototypical-settings}
\end{equation}

The nudge still derives from the MSE cost function as defined in Eq.~\ref{eq:mse-cost-function}. The system also sequentially settles to two fixed steady states $s_{*}$ and $s_{*}^{\beta}$ at the end of  the free and the nudged phase respectively.

\subsubsection{Primitive Function}
We define the primitive function as the function from which the dynamics could derive:
\begin{equation}
    s^{t+1} \approx \frac{\partial\Phi}{\partial s} 
    \label{eq:dynamics-approx}
\end{equation}

which gives, ignoring the activation function $\rho$:
\begin{equation}
    \Phi = \frac{1}{2}s^{T}Ws
\end{equation}

\subsubsection{Learning Rule}
Similarly to the energy-based settings, we now compute the gradient of the primitive function with regard to a parameter of the system in order to perform optimization. The learning rule, expressed as a function of the two equilibrium points $s_{*}$ and $s_{*}^{\beta}$, now reads:

\begin{equation}
\Delta \theta = \frac{1}{\beta}\left(\frac{\partial \Phi}{\partial \theta}(x, s^\beta_*, \theta) - \frac{\partial \Phi}{\partial \theta}(x, s_*, \theta)\right)
\label{eq:annexe-ep-learning-rule-proto}
\end{equation}

The learning rules for the weights and the biases read: 
\begin{equation}
\left\{
    \begin{array}{l}
    \Delta W_{ij}=\frac{1}{\beta}(s_{i,*}^{\beta}s_{j,*}^{\beta}-s_{i,*}s_{j,*})\\
    \Delta b_{i}=\frac{1}{\beta}(s_{i,*}^{\beta}-s_{i,*})
        \end{array}
    \right.
    \label{eq-learning-rule-proto-EP}
\end{equation}

\section{Training Convolutional Networks with Equilibrium Propagation}
\label{annexe:ep-conv-settings}
In this section, we describe and define all operations used to train with EP a convolutional neural network recurrently connected with symmetric synapses. We describe the dynamics and the underlying learning rules for the weights and the biases in the prototypical and the energy-based settings. We denote $N^{\rm conv}$ and $N^{\rm fc}$ the number of convolutional layers and fully connected layers in the convolutional system, and $N^{\rm tot}=N^{\rm conv} + N^{\rm fc}$. The units of the system are denoted by $s$ and listed from $s^{0}=x$ the input to the output $s^{N^{\rm tot}}$.

\subsection{Operations involved in the convolutional system}

We detail here the operations involved in the dynamics of a convolutional RNN in both the prototypical and the energy-based settings.

\begin{itemize}
    \item The 2-D convolution between $w$ with dimension $(C_{\rm in}, C_{\rm out}, F,F)$ and an input $x$ of dimensions $(C_{\rm in}, H_{\rm in}, S_{\rm in})$ and stride one is a tensor $y$ of size $(C_{\rm out}, H_{\rm out}, W_{\rm out})$ defined by:
    \begin{equation}
        y_{c,h,s} = (w \star x)_{c,h,s} = B_c + \sum_{i=0}^{C_{\rm in}-1} \sum_{j=0}^{F-1}\sum_{k=0}^{F-1} w_{c,i,j,k}x_{i,j+h,k+s}, 
    \end{equation}
    where $B_c$ is a channel-wise bias.

    \item The 2-D transpose convolution of $y$ by $\tilde{w}$ is then defined in this work as the gradient of the 2-D convolution with respect to its input:
    \begin{equation}
        (\tilde{w} \star y) = \frac{\partial (w \star x)}{\partial x}\cdot y
    \end{equation}

    \item The dot product ``$\bullet$'' generalized to pairs of tensors of same shape $(C,H,S)$ writes:
    \begin{equation}
        a \bullet b = \sum_{c=0}^{C-1}\sum_{h=0}^{H-1}\sum_{w=0}^{S-1} a_{c,h,s}b_{c,h,s}.
    \end{equation}

    \item The pooling operation $\mathcal{P}$ with stride $F$ and filter size $F$ of $x$:
    \begin{equation}
        \mathcal{P}_{F}(x)_{c,h,s} = \underset{i,j \in [0,F-1]}{\rm max}  \left\{ x_{c, F(h-1)+1+i, F(s-1)+1+j} \right\},
    \end{equation}
    with relative indices of maximums within each pooling zone given by:
    \begin{equation}
        {\rm ind}_{\mathcal{P}}(x)_{c,h,s} = \underset{i,j \in [0,F-1]}{\rm argmax}  \left\{ x_{c, F(h-1)+1+i, F(s-1)+1+j} \right\} = (i^{*}(x,h), j^{*}(x,s)).
    \end{equation}
    \item The unpooling operation $\mathcal{P}^{-1}$ of $y$ with indices ${\rm ind}_{\mathcal{P}}(x)$ is then defined as:
    \begin{equation}
        \mathcal{P}^{-1}(y, {\rm ind}_{\mathcal{P}}(x))_{c,h,s} = \sum_{i,j} y_{c,i,j}\cdot \delta_{h, F(i-1)+1+i^{*}(x,h)} \cdot \delta_{s F(j-1)+1+j^{*}(x,s)},
    \end{equation}
    which consists in filling a tensor with the same dimensions as $x$  with the values of $y$ at the indices ${\rm ind}_{\mathcal{P}}(x)$, and zeroes elsewhere. For notational convenience, we omit to write explicitly the dependence on the indices except when appropriate. We can also see unpooling as the gradient of the pooling operation with respect to its input. 
    \item The flattening operation $\mathcal{F}$ is defined as reshaping a tensor of dimensions $(C,H,S)$ to $(1, CHS)$. We denote by $\mathcal{F}^{-1}$ its inverse.
\end{itemize}

\subsection{Prototypical Settings}

\subsubsection{Equations of the dynamics}

We derive here the dynamics of the convolutional network with symmetric connections and with the mean square error as loss function in the prototypical settings. In this case, the dynamics reads:

\begin{align}
\label{eq:conv-archi-sym}
\left\{
\begin{array}{l}
\displaystyle s^{n+1}_{t+1} = \rho \left( \mathcal{P}(w_{n+1} \star s^{n}_{t}) + \tilde{w}_{n+2} \star \mathcal{P}^{-1}(s^{n+2}_{t}) \right), \qquad \forall n \in [0, N^{\rm conv}-2] \\
\displaystyle s^{N^{\rm conv}}_{t+1} = \rho \left( \mathcal{P}(w_{N^{\rm conv}} \star s^{N^{\rm conv}-1}_{t}) + \mathcal{F}^{-1}({w_{N^{\rm conv}+1}}^{\top} \cdot s^{N^{\rm conv}+1}_{t}) \right), \\
\displaystyle s^{N^{\rm conv} + 1}_{t+1} = \rho \left( w_{N^{\rm conv} + 1} \cdot \mathcal{F}(s^{N^{\rm conv}}_{t}) + {w_{N^{\rm conv} + 2}}^{\top} \cdot s^{N^{\rm conv} + 2}_{t} \right), \\
\displaystyle s_{t+1}^{n+1} = \rho \left( w_{n+1} \cdot s^{n}_{t} + {w_{n+2}}^{\top} \cdot s^{n+2}_{t} \right), \qquad \forall n \in [N^{\rm conv} + 1, N^{\rm tot}-2] \\
\displaystyle s_{t+1}^{N^{\rm tot}} = \rho \left( w_{N^{\rm tot}} \cdot s^{N^{\rm tot}-1}_{t}\right) + \beta(\text{y} - s^{N^{\rm tot}_{t}}), \quad \text{with $\beta=0$ during the first phase,}
\end{array}
\right.
\end{align}
where we take the convention $s^{0}=x$, the input. In this case, we have $s^{N^{\rm tot}}=\hat{y}$, the output layer.
Considering the function:

\begin{align*}
    \Phi(x, s^{1}, \cdots, s^{N^{\rm tot}}) &= 
    \sum_{n= N_{\rm conv} + 2}^{N_{\rm tot} - 1} s^{{n + 1}^\top}\cdot w_{n}\cdot s^{n}
    + s^{N_{\rm conv} + 1}\cdot w_{N_{\rm conv} + 1}\cdot \mathcal{F}(s^{N_{\rm conv}})\\
    &+\sum_{n = 1}^{N_{\rm conv} - 1} s^{n + 1}\bullet\mathcal{P}\left(w_{n + 1}\star s^{n}\right) + s^{1}\bullet\mathcal{P}\left(w_{1}\star x\right), 
\end{align*}

when ignoring the activation function, we have:

\begin{equation}
\label{eq:dphids-SE}
\forall n \in [1, N^{\rm tot}]: \quad s_t^n \approx  \frac{\partial \Phi}{\partial s^n}.
\end{equation}

\subsubsection{Learning rules} 

We derive the learning from the primitive function with the help of Eq. \ref{eq:annexe-ep-learning-rule-proto}. In the prototypical settings, the learning rules read:

\begin{align}
   \left\{
\begin{array}{l}
\Delta w_1  =  \frac{1}{\beta} \left(\mathcal{P}^{-1}(s^{1, \beta}_{*})\star x - \mathcal{P}^{-1}(s^{1}_{*})\star x \right)\\
\forall n \in [1, N_{\rm conv} - 1]:\quad \Delta w_{n + 1}  =  \frac{1}{\beta} \left(\mathcal{P}^{-1}(s^{n + 1, \beta}_{*})\star s^{n, \beta}_{*} - \mathcal{P}^{-1}(s^{n + 1}_{*})\star s^{n}_{*} \right)\\
\Delta w_{N_{\rm conv}+ 1}  =
\frac{1}{\beta}\left(s_{*}^{N_{\rm conv} + 1, \beta}\cdot \mathcal{F}\left(s^{N_{\rm conv}, \beta}_{*}\right)^\top - s_{*}^{N_{\rm conv} + 1}\cdot \mathcal{F}\left(s^{N_{\rm conv}}_{*}\right)^\top \right) \\
\forall n \in [N_{\rm conv} + 2, N_{\rm tot} - 1]: \quad \Delta w_{n}  =
\frac{1}{\beta}\left(s_{*}^{n + 1, \beta}\cdot s_{*}^{{n, \beta}^\top} - s_{*}^{n + 1}\cdot s_{*}^{{n}^\top}  \right) \end{array}
\right. 
\label{deltaconv-sym-SE}
\end{align}

\subsection{Energy-Based Settings}
\subsubsection{Equations of the Dynamics}

Inspired by the primitive function derived in the prototypical settings we define an energy function which applies to an energy-based convolutional system, and rely on the same operations defined above:

\begin{align*}
    E(x, s^{1}, \cdots, s^{N^{\rm tot}}) &= 
    \frac{1}{2}\sum_{n=1}^{N_{tot}}(s^{n})^{2} - \sum_{n=1}^{N_{\rm tot}}b_{n}\rho(s^{n}) - \frac{1}{2}\sum_{n=N_{\rm conv}+2}^{N_{tot}-1}\rho(s^{n+1})^{T}.w_{n}.\rho(s^{n}) \\
    &-\rho(s^{N_{\rm conv} + 1})\cdot w_{N_{\rm conv} + 1}\cdot \mathcal{F}(\rho(s^{N_{\rm conv}}))
    -\sum_{n = 1}^{N_{\rm conv} - 1} \rho(s^{n + 1})\bullet\mathcal{P}\left(w_{n + 1}\star \rho(s^{n})\right) - \rho(s^{1})\bullet\mathcal{P}\left(w_{1}\star x\right) 
\end{align*}

The dynamics is then derived from this energy function with the help of Eq. \ref{eq:def-eb}:

\begin{align}
\label{eq:conv-archi-eb-ep}
\left\{
\begin{array}{l}
\displaystyle \frac{\partial s^{1}}{\partial t} = -s^{1} + \frac{\partial \rho(s^{1})}{\partial s^{1}} \times \left( \mathcal{P}(w_{1} \star \rho(x)) + \tilde{w}_{2} \star \mathcal{P}^{-1}(\rho(s^{2})) \right),\\
\\
\displaystyle \frac{\partial s^{n+1}}{\partial t} = -s^{n+1} + \frac{\partial \rho(s^{n+1})}{\partial s^{n+1}} \times \left( \mathcal{P}(w_{n+1} \star \rho(s^{n})) + \tilde{w}_{n+2} \star \mathcal{P}^{-1}(\rho(s^{n+2})) \right), \qquad \forall n \in [1, N^{\rm conv}-2] \\
\\
\displaystyle \frac{\partial s^{N^{\rm conv}}}{\partial t} = -s^{N^{\rm conv}} + \frac{\partial \rho(s^{N^{\rm conv}})}{\partial s^{N^{\rm conv}}} \times \left( \mathcal{P}(w_{N^{\rm conv}} \star \rho(s^{N^{\rm conv}-1})) + \mathcal{F}^{-1}({w_{N^{\rm conv}+1}}^{\top} \cdot \rho(s^{N^{\rm conv}+1})) \right),\\
\\
\displaystyle \frac{\partial s^{N^{\rm conv} + 1}}{\partial t} = -s^{N^{\rm conv} + 1} + \frac{\partial \rho(s^{N^{\rm conv} + 1})}{\partial s^{N^{\rm conv} + 1}} \times \left( w_{N^{\rm conv} + 1} \cdot \mathcal{F}(\rho(s^{N^{\rm conv}})) + {w_{N^{\rm conv} + 2}}^{\top} \cdot \rho(s^{N^{\rm conv} + 2}) \right), \\
\\
\displaystyle \frac{\partial s^{n+1}}{\partial t} = -s^{n+1} + \frac{\partial \rho(s^{n+1})}{\partial s^{n+1}} \times \left( w_{n+1} \cdot \rho(s^{n}) + {w_{n+2}}^{\top} \cdot \rho(s^{n+2}) \right), \qquad \forall n \in [N^{\rm conv} + 1, N^{\rm tot}-2]\\
\\
\displaystyle \frac{\partial s^{N^{\rm tot}}}{\partial t} = -s^{N^{\rm tot}} + \frac{\partial \rho(s^{N^{\rm tot}})}{\partial s^{N^{\rm tot}}} \times \left( w_{N^{\rm tot}} \cdot \rho(s^{N^{\rm tot}-1})\right) + \beta(\text{y} - s^{N^{\rm tot}}), \quad \text{with $\beta=0$ during the first phase.}\\
\end{array}
\right.
\end{align}

where again we have $s^{N^{\rm tot}}=\hat{y}$, the output layer.

\subsubsection{Learning Rules}

We derive the learning from the primitive function with the help of Eq. \ref{eq:annexe-ep-learning-rule}. In the energy-based settings, the learning rules read:

\begin{align}
   \left\{
\begin{array}{l}
\Delta w_1  =  \frac{1}{\beta} \left(\mathcal{P}^{-1}(\rho(s^{1, \beta}_{*}))\star x - \mathcal{P}^{-1}(\rho(s^{1}_{*}))\star x \right)\\ 
\forall n \in [1, N_{\rm conv} - 1]:\quad \Delta w_{n + 1}  =  \frac{1}{\beta} \left(\mathcal{P}^{-1}(\rho(s^{n + 1, \beta}_{*}))\star \rho(s^{n, \beta}_{*}) - \mathcal{P}^{-1}(\rho(s^{n + 1}_{*}))\star \rho(s^{n}_{*}) \right)\\
\Delta w_{N_{\rm conv}+ 1}  =
\frac{1}{\beta}\left(\rho(s_{*}^{N_{\rm conv} + 1, \beta})\cdot \mathcal{F}\left(\rho(s^{N_{\rm conv}, \beta}_{*})\right)^\top - \rho(s_{*}^{N_{\rm conv} + 1})\cdot \mathcal{F}\left(\rho(s^{N_{\rm conv}}_{*})\right)^\top \right) \\
\forall n \in [N_{\rm conv} + 2, N_{\rm tot} - 1]: \quad \Delta w_{n}  =
\frac{1}{\beta}\left(\rho(s_{*}^{n + 1, \beta})\cdot \rho(s_{*}^{{n, \beta}^\top}) - \rho(s_{*}^{n + 1})\cdot \rho(s_{*}^{{n}^\top})  \right)
\end{array}, 
\right. 
\label{deltaconv-eb}
\end{align}

One should notice that we only need to store the activation $\rho(s)$ of the neurons to compute the gradient for each parameter which turns out to be very interesting when the activation function $\rho$ outputs binary values, as we do in Section \ref{sec:bin-eqprop}.

\section{A Scaling Factor for Equilibrium Propagation}
\label{annexe:scaling-factor}

In this section, we discuss in detail the scaling factor introduced in Section \ref{sec:bin-W-eqprop}. We first describe the initialization of the scaling factor. We then show that a naive initialization for the scaling factor inspired by XNOR-Net leads to good performance, but that tuning more precisely the scaling factor can increase the accuracy. Finally we derive learning rules for the scaling factors allowing EP to optimize by itself the value of the scaling factors. We show that systems learning their scaling factors better fit the training set but also learn faster.

\subsection{Fixed Scaling Factor}

\subsubsection{Initializing $\alpha$ value}
\label{init-alpha-xnor}

Rastegari \etal \cite{DBLP:journals/corr/RastegariORF16} introduced a scaling factor to normalize the binary weights in convolutional architectures with real-valued activations. They obtained the value of the scaling factors by minimizing layer-wise the squared difference between the binary weight vector $\mathbf{B}$ and the real-valued weight vector $\mathbf{W}$, where $\mathbf{B}$ and $\mathbf{W}$ are vectors in $\mathbb{R}^{n}$ and $n=c\times f_{1} \times f_{2}$ with $c$ the number of input channels, and $f_{1}$ and $f_{2}$ the sizes of the filter kernel. The factor $\alpha$ scales $\mathbf{B}$ in the following way:

\begin{equation}
    \mathbf{W}=\alpha\mathbf{B}
    \label{eq:alpha-min}
\end{equation}

They found that the optimal solution is given by:

\begin{equation}
    \alpha^{*} = \frac{||\mathbf{W}||_{l1}}{{\rm dim}(W)}
    \label{eq:alpha-xnor}
\end{equation}

In \cite{DBLP:journals/corr/RastegariORF16}, the real-valued weights $\mathbf{W}$ are updated after each backward pas, and the scaling factor $\alpha$ is re-computed at each forward pass.

For training systems with binary synapses through EP, we first use a scaling factor fixed at initialization. We describe in Alg. \ref{alg:init-alpha} how we initialize the binary weights and the corresponding scaling factors layer-wise:

\begin{algorithm}[H]{\emph{Input}: Architecture: \{${\rm N_{Layers}}$, ${\rm N_{Neurons per layer}}$\}.  \\
\emph{Output}: System having binary weights ($W^{b}$) and scaling factors ($\alpha$) initialized}
    \caption{Initialize the scaling factors ($\alpha$) layer-wise}
    \label{alg:init-alpha}
    \begin{algorithmic}
        \FOR{each Layer}
        \STATE $w^{init}$ = rand(N)  || $w^{init}$ is a random full-precision matrix\\
        \STATE $\alpha = \frac{||w^{init}||_{l1}}{{\rm dim}(w^{init})}$\\
        \STATE $W^{b}=\alpha\times {\rm Sign}(w^{init})$\\
        \ENDFOR
    \end{algorithmic}
\end{algorithm}

The rand() function used in Alg. \ref{alg:init-alpha} stands for the native random initialization of Pytorch which is the Kaiming initialization \cite{DBLP:journals/corr/HeZR015}.

\subsubsection{Naive initialization vs. our initialization}

\begin{figure}[h]
\begin{center}
   \includegraphics[width=1\textwidth]{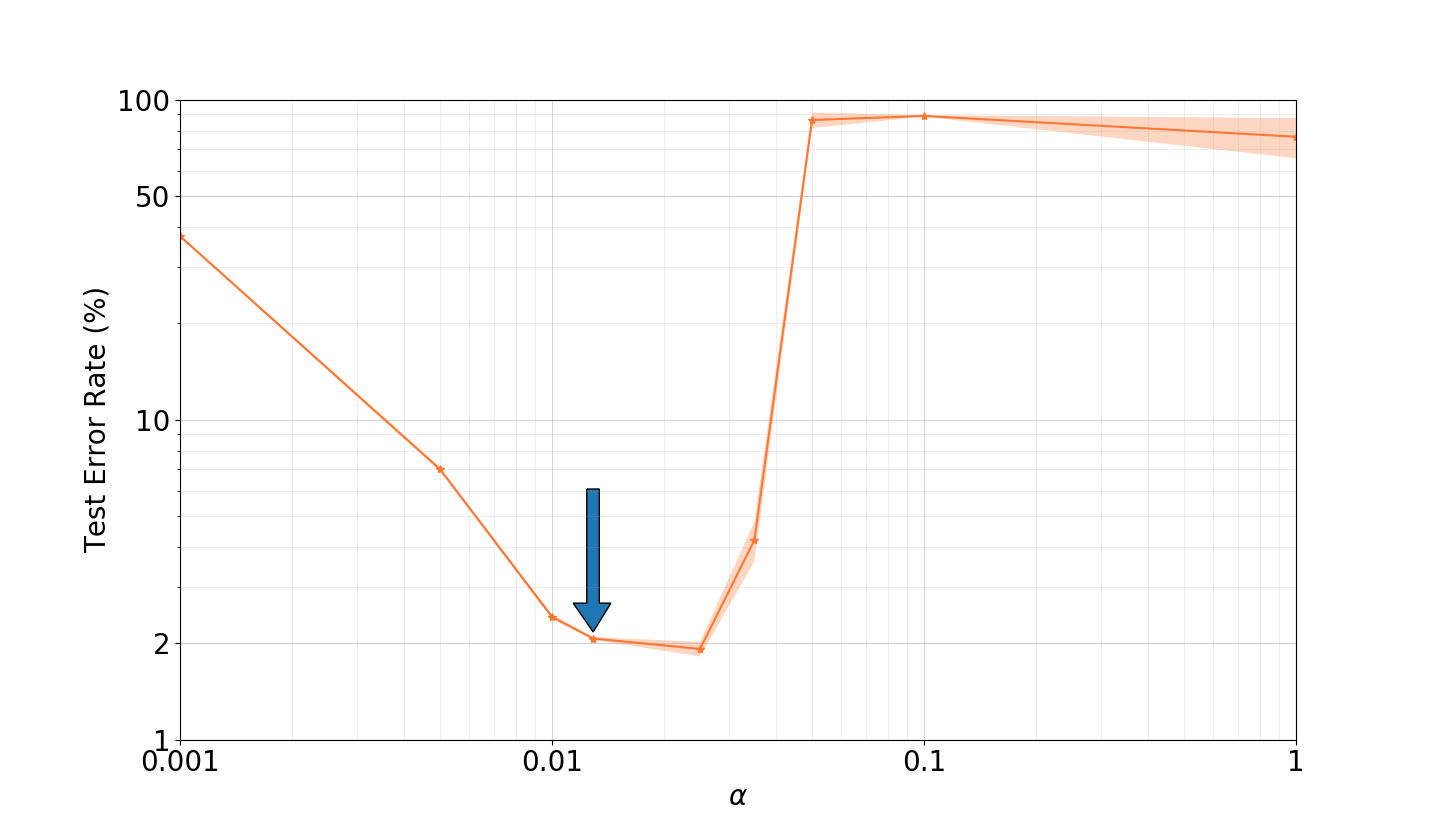}

  \caption{Mean test error $\pm$ standard deviation (computed with 3 trials each) of a 1 hidden layer fully connected neural network on MNIST as a function of the scaling factor $\alpha$ - All dots represent the test error of training performed with $\alpha$ being arbitrarily chosen - $\alpha$ initialized by the method described in \ref{init-alpha-xnor} is indicated by the blue arrow.}
  \label{fig:test_error_vs_alpha}
  \end{center}
\end{figure}

We found that the use of scaling factors $\alpha$ as initialized with Alg. \ref{alg:init-alpha} is crucial to ensure successful training. In fact, if the synapses are initialized to low or too large, we face the vanishing gradient issue as the activation saturate at both 0 or 1.  In order to show this effect we trained a system with a fully connected architecture comprising 1 hidden layer of 4096 neurons, with  binary weights and full-precision activations, for 50 epochs on MNIST. We plot in Fig. \ref{fig:test_error_vs_alpha} the test error obtained with Alg. \ref{alg:init-alpha} for different values of the fixed scaling factors The blue arrow indicate the value corresponding to an initialization of $\alpha$ with Alg. \ref{alg:init-alpha}
(we took the averaged value of both scaling factors in the network to obtain a point on the plot).

We see in Fig. \ref{fig:test_error_vs_alpha} that the test error is highly dependent on the value of the scaling factor. The figure shows that for values of $\alpha$ between about 0.012 and 0.025  the test error is at EP literature level. But even in this range it is not obvious to find the best value for $\alpha$. Moreover, we found that choosing arbitrarily the value of $\alpha$ in deep architectures (fully connected architecture with 2 hidden layers or convolutional architectures) fails at ensuring successful training. We finally chose to initialize the scaling factors with the method of Alg. \ref{alg:init-alpha} as this method is architecture-agnostic and reduces the number of hyperparameters as we already have some to tune. 

In the next section, we address the difficulty to select the best value of $\alpha$ in order to get the best testing accuracy by directly learning the scaling factor with the help of EP.

\subsection{Learning the Scaling Factor with EP}
 
Results in the previous subsection show that optimizing the value of  $\alpha$ can give rise to enhanced performance. Here we show that this optimization can be achieved through EP.
In the context of EP we can indeed derive a learning rule for any parameter in the primitive or energy function. In this section, the scaling factor is first initialized with the method described in Alg. \ref{alg:init-alpha} and is then optimized with SGD with the gradient extracted by EP. For clarity, we decompose the binary weights $W$ from $\pm \alpha$ to $\alpha\times w$ where $w=\pm1$.

\subsubsection{Learning Rules in the Prototypical settings}
\paragraph{Fully connected layers architecture.}
~\\
For a given fully connected layer, the scaling factor $\alpha$ can be introduced in the primitive function of the system as:

\begin{equation}
    \Phi(s) = \frac{1}{2}\alpha\times s^{T}ws
    \label{eq:primitive-function-annexe-alpha-detailled}
\end{equation}

Eq. \ref{eq:annexe-ep-learning-rule-proto} then indicates that the learning rule for the scaling factors in a fully connected architecture in the prototypical settings of EP is:

\begin{equation}
    \Delta\alpha_{l,l+1}=\frac{1}{2\beta}\left( (s_{l}^{T}ws_{l+1})^{\beta}-(s_{l}^{T}ws_{l+1}))^{0}\right) 
    \label{eq:learning-rule-proto}
\end{equation}

where $l$ denotes the index of a layer in the system.

\textbf{Convolutional architecture:}

The scaling factors in use for the classifier are updated with the gradient given by the learning rule stated above.

For convolutional layers, we use one scaling factor per output feature map which gives ${C_{out}}$ scaling factors for a layer with $C_{out}$ feature maps.

Thus for each channel in a convolutional layer $c$ in $C_{out}$ we can write:
\begin{equation}
    \mathcal{P}\left(W_{n + 1}\star s^{n}\right)_{c}=\alpha_{c}\times\mathcal{P}\left(w_{n + 1}\star s^{n}\right)_{c}
    \label{eq:alpha-conv-operation}
\end{equation}

where $W_{n+1}=\alpha_{c}\times w_{n+1}$ are the normalized weights for a channel and $w_{n+1}\in\{-1.1\}$. Following this observation, we can also rewrite a primitive function with $\alpha$ as we did for the fully connected architecture. From this primitive function, we can derive the learning rule for the scaling factors of the convolutional part which reads, channel-wisely:

\begin{align}
   \left\{
\begin{array}{l}
\forall n \in [1, N_{\rm conv} - 1]:\quad \Delta \alpha^{n+1}_{c}  =\frac{1}{\beta}((s^{n + 1}_{c}\bullet\mathcal{P}\left(w_{n + 1}\star s^{n}\right)_{c})^{\beta}-(s^{n + 1}_{c}\bullet\mathcal{P}\left(w_{n + 1}\star s^{n}\right)_{c})^{0}) \\
\Delta \alpha^{1}_{c}=\frac{1}{\beta}((s^{1}_{c}\bullet\mathcal{P}\left(w_{1}\star x\right)_{c})^{\beta}-(s^{1}_{c}\bullet\mathcal{P}\left(w_{1}\star x\right)_{c})^{0})
\end{array}
\right.
\label{eq:deltaALpha}
\end{align}

\subsubsection{Learning Rules in the Energy-Based Settings}
\textbf{Fully connected layers architecture:}

Similarly to the way we introduced $\alpha$ in the primitive function, we re-write the energy function of a fully connected layers architecture as a function of $\alpha$:

\begin{equation}
    E(s)=\frac{1}{2}\sum_{i}s_{i}-\frac{1}{2}\sum_{i\ne j}\alpha_{ij}w_{ij}\rho(s_{i})\rho(s_{i})-\sum_{i}b_{i}\rho(s_{i})
    \label{eq:energy-function-annexe-alpha-detailled}
\end{equation}

Again, with the help of Eq. \ref{eq:annexe-ep-learning-rule} we derive a learning rule for the scaling factors in a fully connected  architecture in the energy-based settings of EP which reads as follow:

\begin{equation}
    \Delta\alpha_{l,l+1}=\frac{1}{2\beta}\left((\rho(s_{l}^{T})w\rho(s_{l+1}))^{\beta}-(\rho(s_{l}^{T})w\rho(s_{l+1}))^{0}\right)
    \label{eq:learning-rule-eb}
\end{equation}

where $l$ denotes the index of a layer in the system.

\textbf{Convolutional architecture:}

The scaling factors in use for the classifier are updated with the gradient given by the learning rule stated above.

In our convolutional networks, we use one scaling factor per feature map which gives ${C_{out}}$ scaling factors for a layer with $C_{out}$ feature maps. For each feature map, we have Eq. \ref{eq:alpha-conv-operation} verified and we can also easily derive the learning rule of the scaling factors of the convolutional layers which reads, channel-wise:

\begin{align}
   \left\{
\begin{array}{l}
\forall n \in [1, N_{\rm conv} - 1]:\quad \Delta \alpha^{n+1}_{c}  =\frac{1}{\beta}((\rho(s^{n + 1}_{c})\bullet\mathcal{P}\left(W_{n + 1}\star \rho(s^{n}\right)_{c}))^{\beta}-(\rho(s^{n + 1}_{c})\bullet\mathcal{P}\left(W_{n + 1}\star \rho(s^{n}\right)_{c})^{0})) \\
\Delta \alpha^{1}_{c}=\frac{1}{\beta}((\rho(s^{1}_{c})\bullet\mathcal{P}\left(W_{1}\star x\right)_{c})^{\beta}-(\rho(s^{1}_{c})\bullet\mathcal{P}\left(W_{1}\star x\right)_{c})^{0})
\end{array}
\right.
\label{eq:deltaALpha-eb}
\end{align}

\subsection{Benefits of Learning the Scaling Factor when the Synapses are Binary and the Activations Full-Precision}

In the next Tables \ref{tab:learn-alpha-epochs-bin-W-mnist-fc}, \ref{tab:learn-alpha-epochs-bin-W-mnist-conv} and \ref{tab:learn-alpha-epochs-bin-W-cifar10-conv}, we report the training and test errors obtained with fixed and dynamical scaling factors on MNIST and CIFAR-10 with different architectures, at mid-training and at the end of the training.

\paragraph{Learning the scaling factor accelerates the training.}

In these tables, we show that the training times are accelerated when the scaling factor is dynamical instead of fixed after initialization. In particular for MNIST, the training is accelerated by a factor over two compared to the fixed scaling, both 
for fully connected and convolutional architectures.

For CIFAR-10 the acceleration is not as large as for MNIST but we struggled to fine-tune the learning rate for the scaling factors and thus better combinations could give larger acceleration.

\paragraph{Systems learning the scaling factors better fit the training set.}

Also in these tables we see that every trainings done with dynamical scaling factors always better fit the training set than trainings done with fixed scaling factors.
We also see in these tables that every training done with dynamical scaling factors better fits the training set than with fixed scaling factors.
Training errors on MNIST are improved by 0.8\% and 0.15\% for fully connected layers architectures having 1 and 2 hidden layers. The convolutional architecture trained on MNIST also gains 0.45\% in terms of training error.
Whereas the fully connected architectures sees the test error also improved alongside the training error, the convolutional architecture sees a slight degradation of the test error due tooverfitting.

The convolutional architecture trained on CIFAR-10 gains 1.1\% training error.

\paragraph{Can learning the scaling factor reduce the memory requirements of the network?}

Finally, learning the scaling allows to train a fully connected architecture with 1 hidden layer of only 512 hidden neurons (the architecture usually trained by EP in the literature) with very low loss of accuracy: +0.2\% testing error and +0.6\% training error (see Table  \ref{tab:learn-alpha-epochs-bin-W-mnist-fc} and Fig. \ref{tab:results-table-bin-W-EP}) whereas with a fixed scaling factor we have +2\% testing error and ~+3\% training error (see Table  \ref{tab:learn-alpha-epochs-bin-W-mnist-fc} and Fig. \ref{fig:error_vs_hiddenNeurons}). However, we did not make this observation across all the architectures that we have studied and only rise it here as a curiosity.

\begin{table*}[!ht]
 \caption{Mean Train and Test errors on MNIST (over 5 trials each) computed after 25 and 50 epochs for two shallow networks with one and two hidden layers with binary synapses trained with EqProp - We denote in the \textit{Learn $\alpha$} column if the scaling factor is learnt or not}
  \centering
  \begin{tabular}{lcccc}
  \toprule
     & & 25 Epochs & 50 Epochs \\
    \hline
    Architecture & Learn $\alpha$ & Test (Train) & Test (Train) \\
    \midrule
    784-4096-10 & \ding{55} & 2.14 (0.92) & 2.07 (0.77) \\
    784-4096-10 & \checkmark & 1.66 (0.03) & \textbf{1.7 (0)} \\
    784-512-10 & \checkmark & 2.45 (1.24) & \textbf{2.2 (0.7)} \\
    784-4096(2)-10 & \ding{55} & 2.47 (0.4) & 2.48 (0.15) \\
    784-4096(2)-10 & \checkmark & 2.27 (0.02) &  \textbf{2.28 (0)} \\
    \bottomrule
  \end{tabular}
  \label{tab:learn-alpha-epochs-bin-W-mnist-fc}
\end{table*}

\begin{table*}[!ht]
 \caption{Mean Train and Test errors on MNIST (over 5 trials each) with a convolution network with binary synapses trained with EqProp - We denote in the \textit{Learn $\alpha$} column if the scaling factor is learnt or not}
  \centering
  \begin{tabular}{lcccc}
  \toprule
     & & 25 Epochs & 50 Epochs \\
    \hline
    Architecture & Learn $\alpha$ & Test (Train) & Test (Train)\\
    \midrule
    1-32-64-(fc) & \ding{55} & 0.92 (0.63) & 0.84 (0.46) \\
    1-32-64-(fc) & \checkmark & 0.79 (0.16) & \textbf{0.88 (0.047)}  \\
    \bottomrule
  \end{tabular}
  \label{tab:learn-alpha-epochs-bin-W-mnist-conv}
\end{table*}

\begin{table*}[!ht]
 \caption{Mean Train and Test errors on CIFAR-10 (over 5 trials each) with a convolution network with binary synapses trained with EqProp - We denote in the \textit{Learn $\alpha$} column if the scaling factor is learnt or not}
  \centering
  \begin{tabular}{lcccc}
  \toprule
     & & 100 Epochs & 500 Epochs \\
    \hline
    Architecture & Learn $\alpha$ & Test (Train) & Test (Train)\\
    \midrule
    3-68-128-256-256-(fc) & \ding{55} & 18.4 (13.4) & 16.8 (6.9) \\
    3-68-128-256-256-(fc) & \checkmark & 17.6 (11.86) & \textbf{15.54 (5.54)} \\
    \bottomrule
  \end{tabular}
  \label{tab:learn-alpha-epochs-bin-W-cifar10-conv}
\end{table*}
 


\section{Propagation of the Error Signal with Binary Activations}
\label{annexe:error-signal}
In this section we discuss how the binary activation function can cancel the error signal flowing from the nudged output neurons to the other upstream layers. We then explain how we can enhance the error signal in order to yield a nudging force strong enough to propagate throughout the system.

\subsection{The Error Signal Flows Into the System if the First Hidden Layer is Sensitive to it}
\label{annexe:first-hidden-sensible}

A layered architecture trained with EP makes the system sensitive to the error signal if and only if neurons between the output layer - where the error signal is applied - and a neuron of interest, are sensitive to the target. The first hidden layer is in this sense a bottleneck for the error signal. It often receives more forward signal from the other hidden layers than backward signal from the output layer which only has 10 neurons for MNIST and CIFAR-10. During the nudging phase of EP, only one neuron in the output layer is nudged to be $1$, the others being nudged to $0$. And this little change in the output layer in not sufficient for the first hidden layer to reach the criteria of good error signal Eq. \ref{eq:condition-bin-neurons}. Therefore, the binary activations of the neurons in the first hidden layer do not change and the error signal is blocked.

Once the first hidden layer changes its binary activations, the error signal can flow through the network. In fact, it has more impact on the others hidden layers because it is often larger than the output layer and matches approximately the size of the others layers thus it is more likely to impact the binary activation of the next hidden layer. Augmenting the error signal is crucial to train systems with binary synapses and activations with EP.   

\begin{figure}[ht!]
\begin{center}
   \includegraphics[width=1\textwidth]{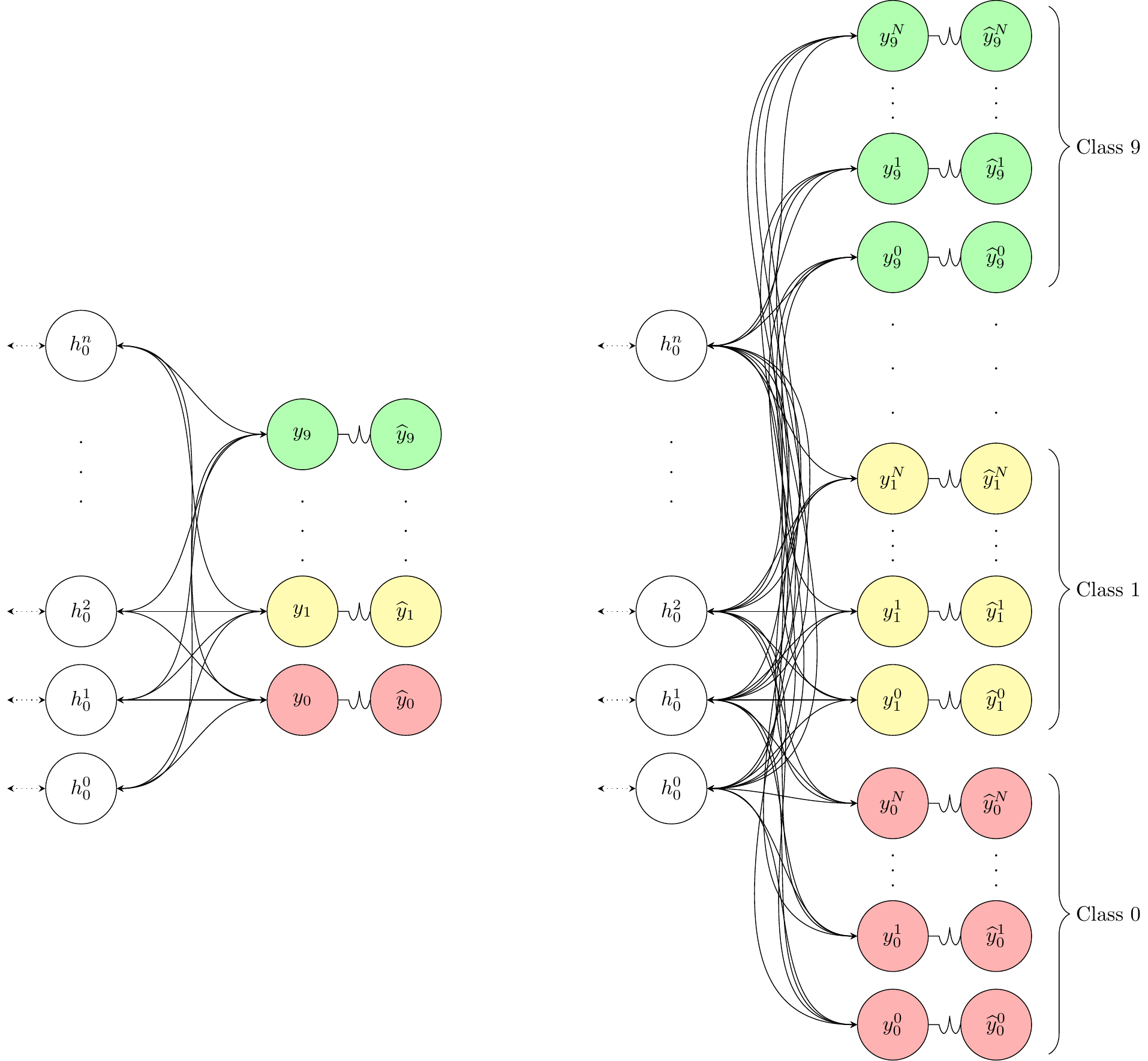}
\end{center}
  \caption{Left: Schematic of the classic output layer with one output neuron per class - compared to Right: the enlarged output layer where we have $N_{\rm perclass}$ output neurons per class - Hidden units are denoted by $h$, output units by $y^{x}$ and the target units by $\widehat y^{x}$ - We represent the nudging of the output neurons by the corresponding target units with the small springs on the schematic - For simplicity we drew this schematic for datasets having 10 output classes but it can be applied to any dataset -  Dashed arrows on the left hand of both networks indicate the bidirectional connections with the rest of the network}
  \label{fig:large-output-layer-schema}
\end{figure}

\subsection{An Enlarged Output Layer for a Greater Error Signal}

The scaling factors introduced in Section \ref{sec:bin-W-eqprop} to normalize the binary weights and not to saturate the activations show limitations with binary activations. Neurons with binary activations can sometimes indeed no longer propagate the error signal.

We enlarged the output layer to solve this issue as described in Fig. \ref{fig:large-output-layer-schema}. This enlargement of the output layer makes the neurons having a binary activation again sensitive to the error signal and their activation can change during the nudging phase. This opens the path to training deeper architectures with binary activations and weights with EP. Our solution is similar in spirit to the augmented output layer used by \cite{bartunov2018assessing}.

\subsection{Making Predictions with an Enlarged Output Layer}
\label{annexe:error-signal-prediction}

Usually an output layer has as many neurons as the number of classes in the dataset and the prediction is the \textit{argmax} of the output layer.

But when we train systems having binary activations the output layer is augmented and it is not straightforward to make a prediction taking the \textit{argmax} of the output layer. We describe here two methods to make a prediction with the enlarged output layer. We used both methods in our simulations and show they give similar accuracy in the end.

\paragraph{Making predictions by averaging each sub-class.} 
This first method allow us to retrieve a situation similar to the classic output layer having one neuron per class. In fact we first average the internal state - or pre-activation - of each neuron belonging to a class which gives 10 averaged values and the prediction is taken as the \textit{argmax} of these averaged values.

\paragraph{Making predictions with one neuron per sub-class.} 

The first method we describe above to make the prediction could reveal to be computationally and time expensive and costly to realize on digital hardware. A second, more hardware-friendly, method is to look at the state of only one output neuron per class and take the argmax of these "single-neurons".

\begin{figure}[ht!]
\begin{center}
   \includegraphics[width=0.8\textwidth]{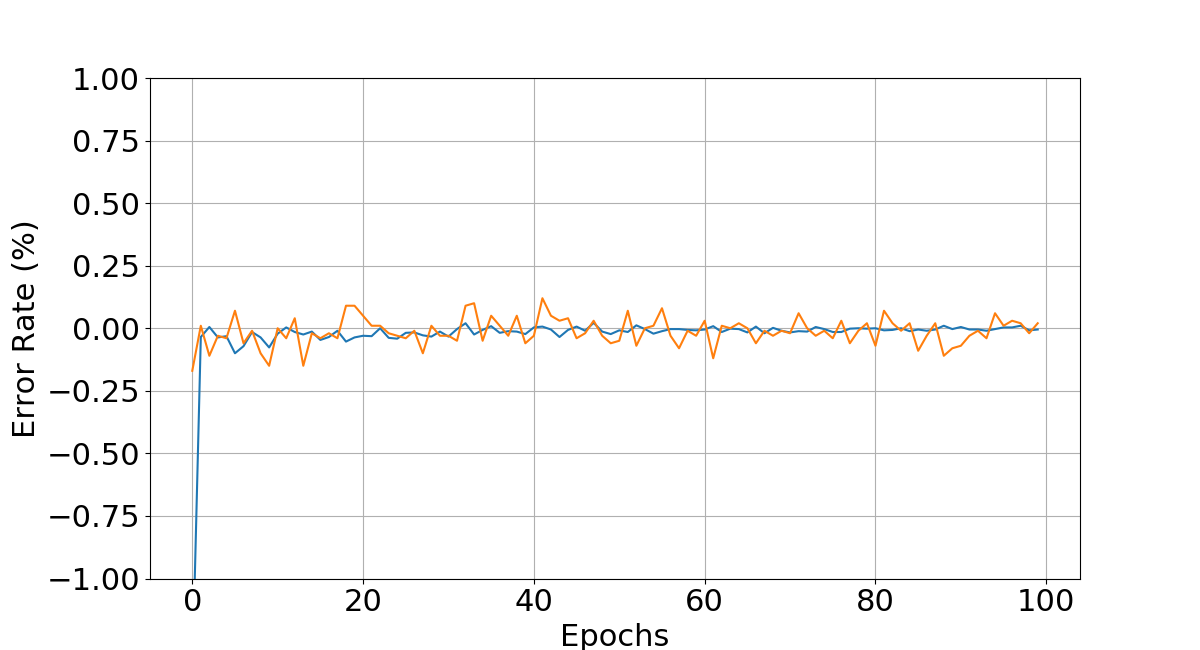}
\end{center}
  \caption{Difference of the train and test errors computed with the averaging method and the method with only one neuron per subclass for a system with one hidden layer of 8192 neurons and 100 output neurons}
  \label{fig:MNIST-1fc-8192-100-diff}
\end{figure}

\begin{figure}[ht!]
\begin{center}
   \includegraphics[width=0.8\textwidth]{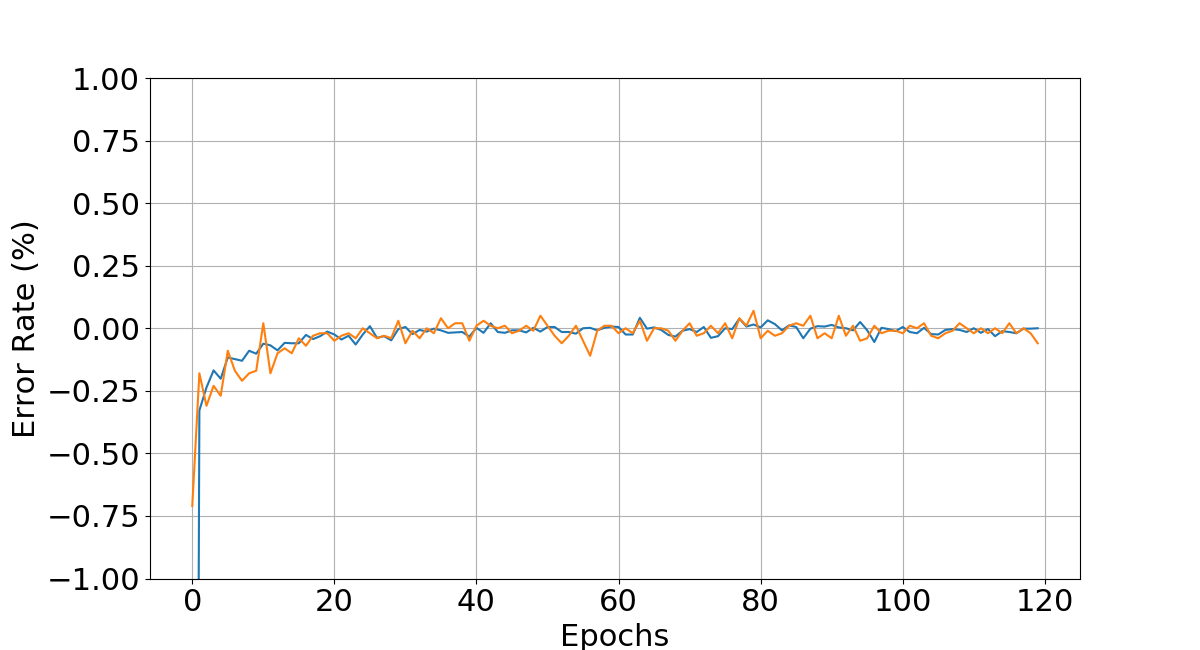}
\end{center}
  \caption{Difference of the train and test errors computed with the averaging method and the method with only one neuron per subclass for a system with two hidden layers of 8192 neurons and 8000 output neurons}
  \label{fig:MNIST-2fc-8192-8000-diff}
\end{figure}

\paragraph{Comparison of the two methods.} 
We want to see if the second method, which constitutes a great simplification of the prediction process, performs as well as the first method. For this purpose, we plot in Fig. \ref{fig:MNIST-1fc-8192-100-diff} and Fig. \ref{fig:MNIST-2fc-8192-8000-diff} the difference of the training and the testing errors of two fully connected architectures with binary synapses and activations and 1 and 2 hidden layers trained on MNIST computed with the two methods described above. We see that for the network with 1 hidden layer, the difference between the two methods does not exceed 0.1\% for both training and testing errors. For the network with 2 hidden layers, despite the fact that the difference starts at a high level with more than 0.7\% of difference for the testing error and more than 1\% for the training error, in the end of the training process the differences have decreased to almost 0\%. Both figures show the effectiveness of the second method at making the predictions at a lower cost both computationally and in time than the first method.

\section{Simulations Details - Hyperparameters and Training Curves}
\label{annexe:simulations-details}

\subsection{Binary Synapses}
\label{subannexe: bin-synapses}
We detail in this section all settings and parameters used for the simulations for EP with binary synapses and full-precision activations (hardsigmoid activation function). We ran the simulations with PyTorch and speed them up on a GPU. The duration of the simulations runs from $30$ mins for the shallow network to $5$ days for the convolutional architecture on CIFAR-10. 

For these simulations, we use the prototypical settings of EP for the sake of saving simulation time. The energy-based settings would perform the same way but such models are much longer to train.

We found that comparatively to full-precision models trained by EP, the error signal vanishes through the system and thus deep layers need a greater learning rate for the biases and greater $\gamma$ for the weights. All hyperparameters are reported in Table \ref{tab:hyper-bin-W-EP}. 

The target is one-hot encoded and the prediction is computed by taking the argmax of the state of the output neurons. The output layer is designed in a way that we have one output neuron per class of the dataset. We initialize the binary weights taking the sign of randomly-initialized weights matrices.

We choose the sign of beta randomly at each mini-batch which is known to give better results \cite{DBLP:journals/corr/ScellierB16,laborieux2020scaling}. For all simulations we used mini-batches of size of $64$ as we found it performs better.

All figures report the mean of the training and testing errors computed with 5 trials each $\pm$ 1 standard deviation.

\paragraph{MNIST - fully connected layer - 1 hidden layer.} 
We train a network with a fully connected architecture and 1 hidden layer on MNIST. We first tuned the EP hyperparameters ($T$, $K$, $\beta$) making EP gradient estimates match those given by BPTT \cite{NIPS2019_8930}. At the same time we tuned BOP hyperparameters in order to fit the flipping metric (Eq. \ref{eq:metric-bop}) in the range leading to successful training as described in Section \ref{sec:bin-W-eqprop}.
We found that contrarily to Helwegen \etal \cite{NEURIPS2019_9ca8c9b0}, the flipping metric starts at high level (between $0$ and $-4/-5$) and decreases over epochs to reach a region below -5.

We initialize one scaling factor per weight matrix with the method described in Alg. \ref{alg:init-alpha}. When the scaling factors are learnt, we use the same learning rate for all scaling factors. Despite the fact that the learning rule for the scaling factors requires the sign of the weights $\pm 1$ for the computation, we found that using the scaled weights $\pm \alpha$ performs the same way so we used the scaled weight matrix to compute the gradient.

\begin{figure}[!ht]
\begin{center}
   \includegraphics[width=0.5\textwidth]{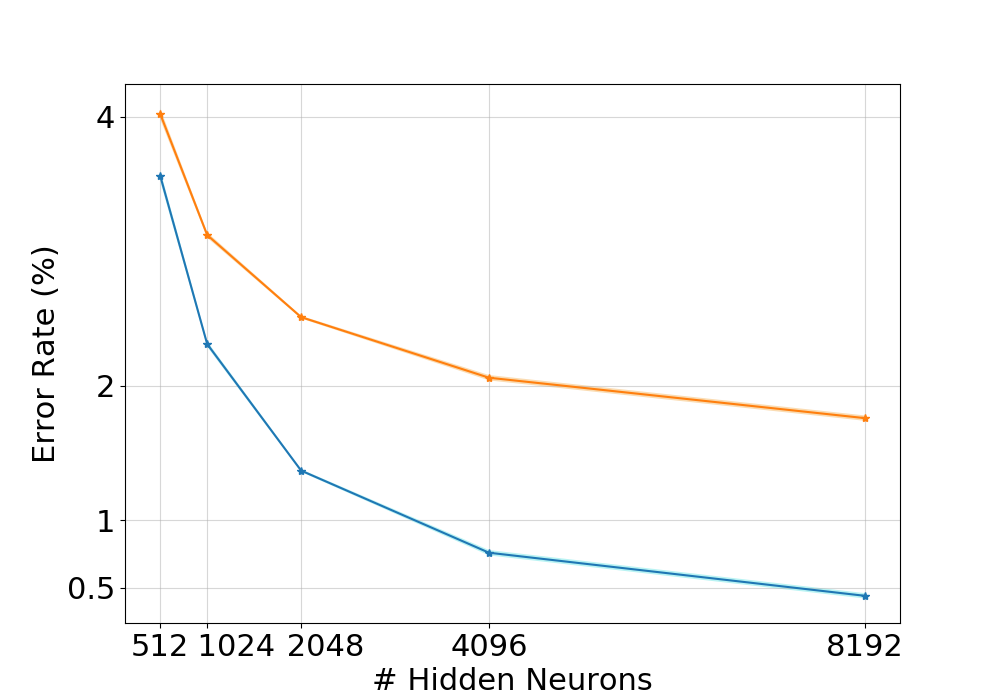}
\end{center}
  \caption{Averaged train (blue) and test (orange) errors on MNIST with a fully connected architecture with one hidden layers as a function of the number of hidden neurons - We average the errors over 5 trials and plot the average $\pm$ 1 standard deviation}
  \label{fig:error_vs_hiddenNeurons}
\end{figure}

To reach an accuracy at levels of reported results in the literature with such architecture trained by EP on MNIST, we needed to increase by 8 the number of neurons in the hidden layer as shown in Fig. \ref{fig:error_vs_hiddenNeurons} when the scaling factors are fixed which justifies the architecture we trained: 784-4096-10.

We report all hyperparameters in Table \ref{tab:hyper-bin-W-EP}. We initialize the biases with the native PyTorch random initialization and the state of the neurons to zero as it has proven to perform better.

\begin{figure}[ht!]
    \begin{minipage}[c]{.46\linewidth}
        \centering
        \includegraphics[width=1\textwidth]{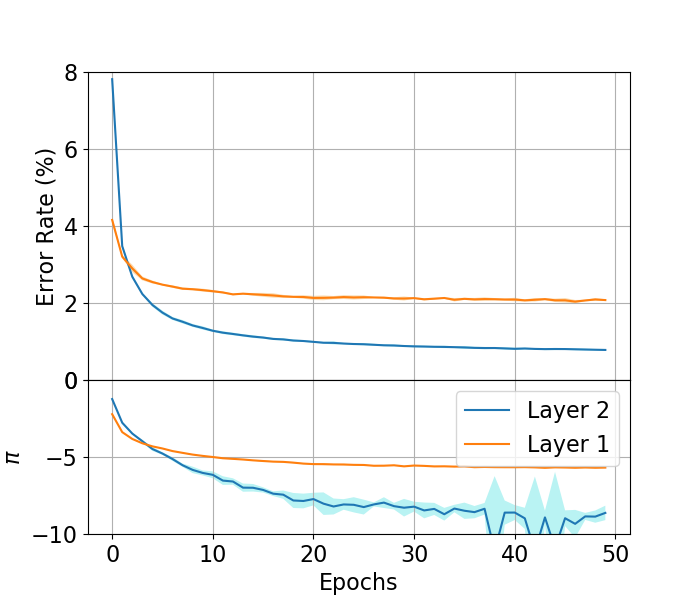}
        \caption{Top: Train (blue) and test (orange) error on MNIST with a fully connected architecture with one hidden layer of 4096 neurons trained with EP with binary synapses - The scaling factors are fixed - Down: metric of weights flipped during an epoch, given for each weights matrix from input to output}
        \label{fig:MNIST-1fc-4096-EP}
    \end{minipage}
    \hfill%
    \begin{minipage}[c]{.46\linewidth}
        \centering
        \includegraphics[width=1\textwidth]{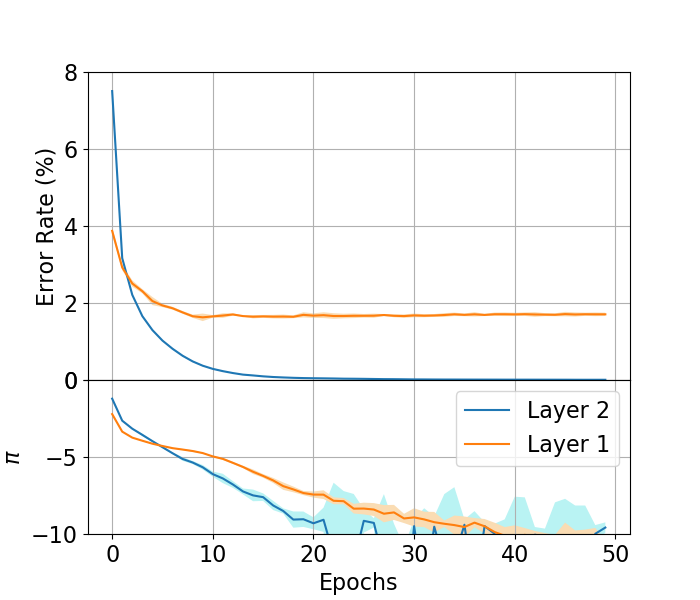}
        \caption{Top: Train (blue) and test (orange) error on MNIST with a fully connected architecture with one hidden layer of 4096 neurons trained with EP with binary synapses - The scaling factors are learnt - Down: metric of weights flipped during an epoch, given for each weights matrix from input to output}
        \label{fig:MNIST-1fc-4096-EP-learnAlpha}
    \end{minipage}
\end{figure}

We performed two sets of simulations: 
\begin{itemize}
    \item Simulations where the scaling factors were fixed which achieve accuracy (Table \ref{tab:results-table-bin-W-EP}, Fig. \ref{fig:MNIST-1fc-4096-EP}) close to those reported in the EP literature: \cite{NIPS2019_8930}.
    \item Simulations where the scaling factors were learnt. We show that learning the scaling factors improves by a considerable margin the training -0.7\% and the testing -0.4\% errors: Fig. \ref{fig:MNIST-1fc-4096-EP}, Fig. \ref{fig:MNIST-1fc-4096-EP-learnAlpha} and Table \ref{tab:learn-alpha-epochs-bin-W-mnist-fc}. We link the better testing error to a better fit on the training set as the network seems to overfit a bit: the testing error starts to increase after 10 epochs which also highlights the fact that when we learn the scaling factors, we can use less neurons per hidden layer and still get accuracy close to those reported in the EP literature. We also report that the training is at least five times faster, as after 10 epochs the training and testing errors are below the levels obtained after 50 epochs with fixed scaling factors. Learning the scaling factors makes the flipping metric of BOP to decrease more quickly than when the scaling factors are fixed.
\end{itemize}

\paragraph{MNIST - fully connected layer - 2 hidden layers.} 
We train a network with a fully connected architecture which has 2 hidden layers on MNIST. We initially chose EP and BOP hyperparameters close to the hyperparameters chosen for training the network with one hidden layer network and then fine-tuned them to achieve the best accuracy. The metric of BOP (Eq. \ref{eq:metric-bop}) also decreases over epochs to reach a level below -5 in the good range for BOP. 

Again, we initialize with Alg. \ref{alg:init-alpha} one scaling factor per weight matrix which gives 3 scaling factors for this architecture. We also use the same learning rate for all scaling factors and the scaled weights for computing the gradient as done with the architecture which has 1 hidden layer.

We kept the same number of neurons (4096) in each hidden layer as for the architecture which has only 1 hidden layer.

We report all hyperparameters in Table \ref{tab:hyper-bin-W-EP}. We initialize the weights with the native PyTorch random initialization and the state of the neurons to zero as it has proven to perform better.

\begin{figure}[h]
    \begin{minipage}[c]{.46\linewidth}
        \centering
        \includegraphics[width=1\textwidth]{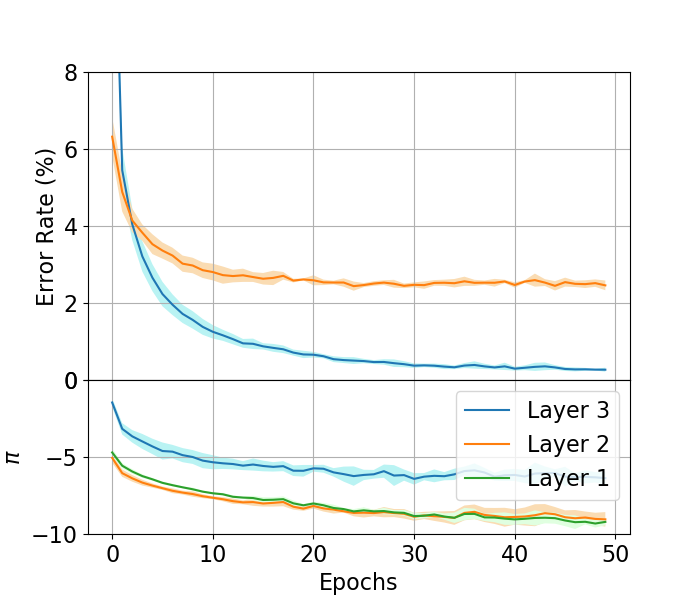}
        \caption{Top: Train (blue) and test (orange) error on MNIST with a fully connected architecture with two hidden layers of 4096 neurons trained with EP with binary synapses - The scaling factors are fixed - Down: metric of weights flipped during an epoch, given for each weights matrix from input to output}
  \label{fig:MNIST-2fc-4096-4096-EP}
    \end{minipage}
    \hfill%
    \begin{minipage}[c]{.46\linewidth}
        \centering
        \includegraphics[width=1\textwidth]{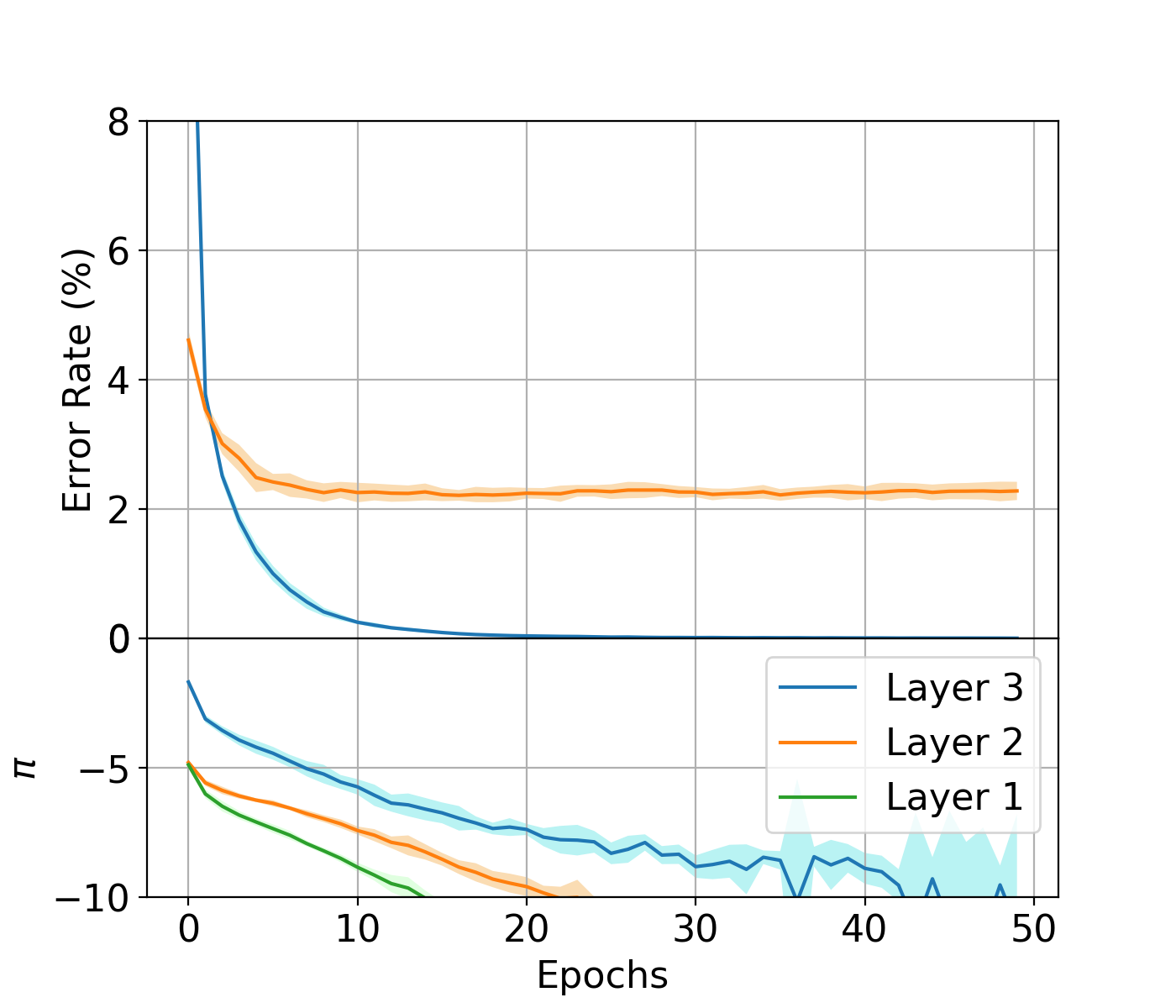}
        \caption{Top: Train (blue) and test (orange) error on MNIST with a fully connected architecture with two hidden layers of 4096 neurons trained with EP with binary synapses - The scaling factors are learnt - Down: metric of weights flipped during an epoch, given for each weights matrix from input to output}
  \label{fig:MNIST-2fc-4096-4096-EP-learnAlpha}
    \end{minipage}
\end{figure}

We performed two sets of simulations: 
\begin{itemize}
    \item Simulations where the scaling factors were fixed, which achieve accuracy (Table \ref{tab:results-table-bin-W-EP}, Fig. \ref{fig:MNIST-2fc-4096-4096-EP}) close to those reported in the EP literature \cite{NIPS2019_8930}. 
    \item Simulations where the scaling factors were learnt. We show that learning the scaling factors improves a lot the fit by 0.15\% on the train set and thus improves the testing accuracy by 0.2\% in Fig. \ref{fig:MNIST-2fc-4096-4096-EP}, Fig. \ref{fig:MNIST-2fc-4096-4096-EP-learnAlpha} and Table \ref{tab:learn-alpha-epochs-bin-W-mnist-fc}. Finally learning the scaling factors also speed up the training by at least a factor 5.
\end{itemize}

\textbf{MNIST - convolutional architecture:}
We train a convolutional network on MNIST. The architecture used consists in the following: 2 convolutional layers of respectively 32 and 64 channels. We use convolutional kernels of size $5\times 5$, padding of 2 and a stride of 1. Each convolutional operation is followed by a $2\time2$ Max Pooling operation with a stride of $2$. We flatten the output of the last convolutional layer to feed the output layer of 10 neurons.

We tuned EP hyperparameters (T,K,$\beta$) making EP gradient estimates match the gradient given by BPTT \cite{NIPS2019_8930}. We tuned BOP hyperparameters to make the metric in the range below -5.

The scaling factors $\alpha$ are initialized channel-wise in each convolutional layer which gives 32 scaling factors for the first convolutional layer and 64 scaling factors for the second convolutional layer with the architecture used here. Again we use the scaled weights to compute the gradient of each scaling factor.

\begin{figure}[ht!]
    \begin{minipage}[c]{.46\linewidth}
        \centering
        \includegraphics[width=1\textwidth]{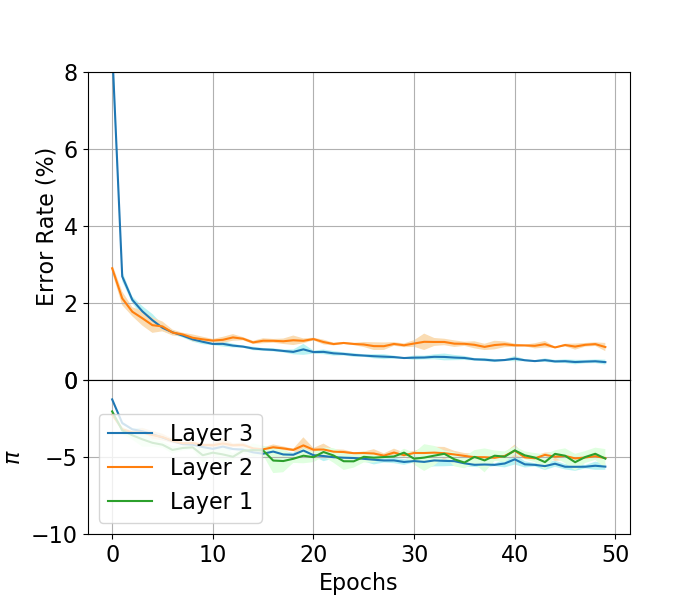}
        \caption{Top: Train (blue) and test (orange) error on MNIST with a convolutional architecture with 2 convolutional layers of respectively 32 and 64 channels trained with EP with binary synapses - The scaling factors are fixed -  Down: metric of weights flipped during an epoch, given for each weights matrix from input to output.}
        \label{fig:MNIST-conv-EP}
    \end{minipage}
    \hfill%
    \begin{minipage}[c]{.46\linewidth}
        \centering
        \includegraphics[width=1\textwidth]{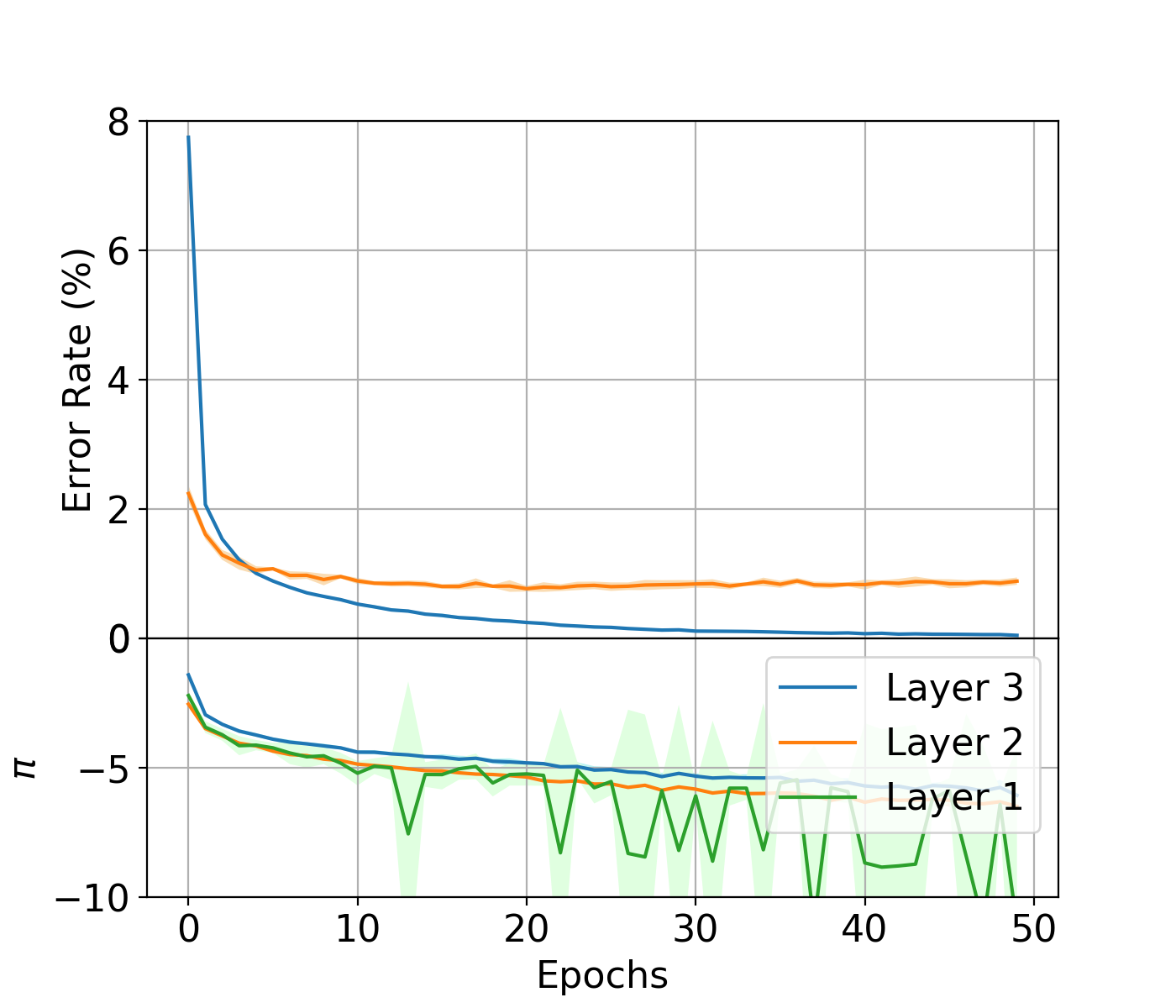}
        \caption{Top: Train (blue) and test (orange) error on MNIST with a convolutional architecture with 2 convolutional layers of respectively 32 and 64 channels trained with EP with binary synapses  - The scaling factors are learnt - Down: metric of weights flipped during an epoch, given for each weights matrix from input to output.}
        \label{fig:MNIST-conv-EP-learnAlpha}
    \end{minipage}
\end{figure}

We performed two sets of simulations:
\begin{itemize}
    \item Simulations where the scaling factors were fixed. We report accuracy slightly below the one reported with EP on MNIST with the same convolutional architecture in \cite{NIPS2019_8930}: -0.4\% for the training and -0.2\% for the testing errors. Two things one: as underscored before, BOP seems to regularize the training with EP but also we used the sign of $\beta$ randomly which is known to better estimate the gradient given by EP and thus improve the training. 
    \item Simulations where the scaling factors were learnt. Learning the scaling factors allow the system to better fit the training set (-0.5\% of training error). But this makes the system to overfit as the testing error increases to 0.88\% after 50 epochs after having reached a minimum at 0.76\% after 25 epochs. Learning the scaling factors also decreases more the flipping metric $\pi$ as shown in Fig. \ref{fig:MNIST-conv-EP}, Fig. \ref{fig:MNIST-conv-EP-learnAlpha} and Table \ref{tab:learn-alpha-epochs-bin-W-mnist-conv}.
\end{itemize}

\begin{figure}[h]
    \begin{minipage}[c]{.46\linewidth}
       \centering
   \includegraphics[width=1\textwidth]{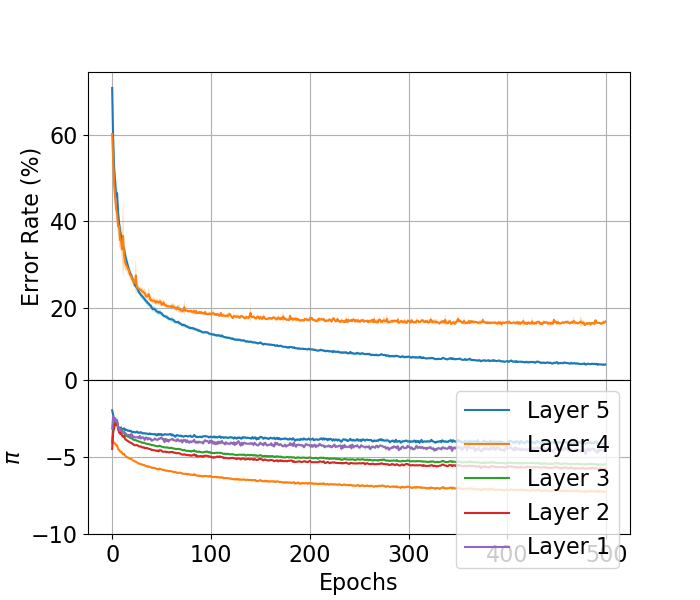}
  \caption{Top: Train (blue) and test (orange) error on CIFAR-10 with a convolutional architecture with 4 convolutional layers of respectively 64, 128, 256 and 256 channels trained with EP with binary synapses - The scaling factors are fixed - Down: metric of weights flipped during an epoch, given for each weights matrix from input to output.}
  \label{fig:CIFAR10-conv-EP}
    \end{minipage}
    \hfill%
    \begin{minipage}[c]{.46\linewidth}
        \centering
        \includegraphics[width=1\textwidth]{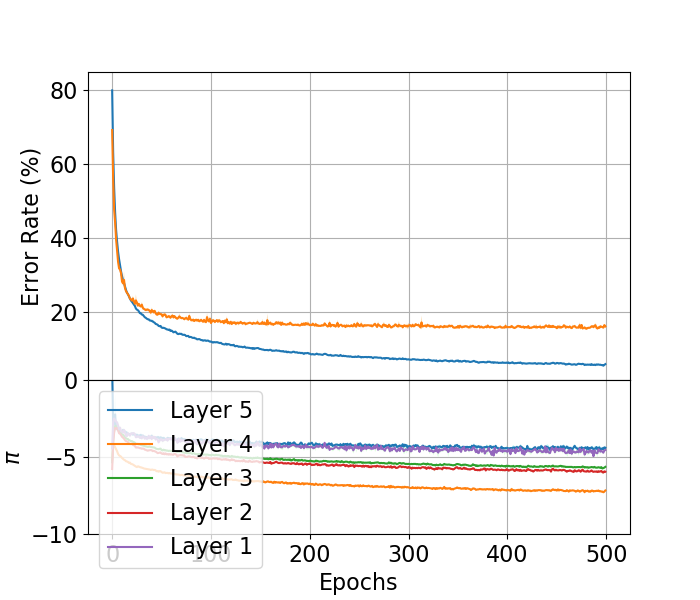}
  \caption{Top: Train (blue) and test (orange) error on CIFAR-10 with a convolutional architecture with 4 convolutional layers of respectively 64, 128, 256 and 256 channels trained with EP with binary synapses - The scaling factors are learnt - Down: metric of weights flipped during an epoch, given for each weights matrix from input to output.}
  \label{fig:CIFAR10-conv-EP-learnAlpha}
    \end{minipage}
\end{figure}

\begin{table*}[!ht]
 \caption{Hyperparameters used for training systems with EP and binary synapses - $lrBias$ are the learning rates used for updating the biases with SGD and given from input to output layer - $\gamma$ is layer-dependent and given from input to output layer}
  \centering
  \begin{tabular}{lllccccccc}
  \toprule
     & & & \multicolumn{3}{c}{EP} & & \multicolumn{2}{c}{BOP} &  \\
    \cline{4-6}
    \cline{8-9}
    Dataset     & Method     & Architecture  & T & K &  $\beta$& & $\gamma$ & $\tau$ &  $lrBias$ \\
    \midrule
    MNIST & fc  & 784-4096-10 & 50 & 10 & 0.3 & & 1e-4-1e-5 & 5e-7 &  0.05-0.025  \\
    MNIST     & fc & 784-4096(2)-10 & 250 & 10 & 0.3 & & 2e-5-2e-5-5e-6 & 5e-7 &  0.2-0.1-0.05   \\
    MNSIT     & conv  & 1-32-64-fc & 150  & 10  & 0.3 & & 5e-8 & 1e-8 & 0.1-0.05-0.025  \\
    CIFAR-10 & conv & 3-64-128-256(2)-fc & 150 & 10 & 0.3 & & 1e-7(2)-2e-7(2)-5e-8 & 1e-8 &  0.4-0.2-0.1-0.05-0.025  \\
    \bottomrule
  \end{tabular}
  \label{tab:hyper-bin-W-EP}
\end{table*}

\paragraph{CIFAR-10 - convolutional architecture.}
We train a convolutional network on CIFAR-10. The architecture used consists in the following: 3-64-128-256-256-fc(10): 4 convolutional layers of respectively 64, 128, 256 and 256 channels, one output layer of 10 neurons. We use convolutional kernels of size $5\times 5$, padding of 2 and a stride of $1$. Each convolutional operation is followed by a $2\time2$ Max Pooling operation with a stride of $2$. We flatten the output of the last convolutional layer to feed the output layer.

Because we used twice as less feature maps at each convolutional layer to speed up our training simulations compared to the original network Laborieux \etal \cite{laborieux2020scaling}, we used the available code \footnote{The code is available at: \href{https://github.com/Laborieux-Axel/Equilibrium-Propagation}{https://github.com/Laborieux-Axel/Equilibrium-Propagation}.} to run simulations with the same architecture as ours to benchmark our technique. 

We chose EP hyperparameters equal to those used for the convolutional architecture trained on MNIST as it has shown to work well. We tuned BOP hyperparameters to make the metric in the good range for BOP.

The scaling factors $\alpha$ are initialized channel-wise in each convolutional layer which for instance gives 64 scaling factors for the first convolutional layer with the architecture used here. Again we use the scaled weights to compute the gradient of each scaling factor.

We performed two sets of simulations:
\begin{itemize}
    \item Simulations where the scaling factors were fixed which achieve accuracy (Table \ref{tab:results-table-bin-W-EP}, Fig. \ref{fig:CIFAR10-conv-EP}) close to those reported in the EP literature (\cite{laborieux2020scaling}).
    \item Simulations where the scaling factors were learnt. We show that learning the scaling factors improves a lot the fit by 1.4\% on the train set and thus improves the testing accuracy by 1.2\% in Fig. \ref{fig:CIFAR10-conv-EP}, Fig. \ref{fig:CIFAR10-conv-EP-learnAlpha} and Table \ref{tab:learn-alpha-epochs-bin-W-cifar10-conv}.
\end{itemize}

We pre-processed CIFAR-10 with the following data augmentation and normalization techniques before feeding it to the system: 
\begin{itemize}
    \item Random Horizontal Flip with $p = 0.5$
    \item Random Crop with $padding = 4 $
    \item Normalization with $\mu=(0.4914, 0.4822, 0.4465)$ and $\sigma = (0.247, 0.243, 0.261)$ for each rgb input channel.
\end{itemize}

\subsection{Binary Synapses and Activations}

We detail in this section all settings and parameters used for the simulations of EP with binary synapses and binary activations. We ran the simulations with PyTorch and sped them up on a GPU. For all simulations we used mini-batches of size of $64$ as we found it performs better. The time of the simulations runs from $30$ mins for the shallow network to $5$ days for the convolutional architecture on CIFAR-10. 

For the simulations of EP with binary synapses and binary activations we use the energy-based settings of EP with the rules derived in Section \ref{sec:bin-eqprop} such as the pseudo-derivative of the Heavyside step function and the enlarged output layer.

In this section, we explore how $\tau$ can be finely tuned layer-wise in order to give the best performance while having the same $\gamma$ for all layers which could be more hardware friendly as we could use the same devices to store the momentum and only change the threshold layer-wise. All hyperparameters are reported in Table \ref{tab:hyper-bin-W-EP}. 

The target is one-hot encoded and then replicated $N_{\rm perclass}$ times to match the size of the output layer. We make the prediction with the two methods described in \ref{annexe:error-signal-prediction}. We initialize the binary weights taking the sign of randomly-initialized weights matrices (native random initialization of Pytorch which is the Uniform Kaiming initialisation).

The input data is kept full-precision thus the MAC operation for the first layer of each architecture is full-precision and also the gradient.

We choose the sign of beta randomly at each mini-batch which is known to give better results \cite{DBLP:journals/corr/ScellierB16}, \cite{laborieux2020scaling} for all simulations except when training the network with the fully connected architecture and which has 2 hidden layers where we only used $\beta>0$.

We used the Heaviside step function as the binary activation function as emphasised in Section \ref{sec:bin-eqprop}. For defining the pseudo-derivative function $\hat\rho^{'}(s)$ (Eq. \ref{eq:def-pseudo-derivative}) we used $\sigma=0.5$ despite using a binary activation.

All figures report the mean of the training and testing errors computed with 5 trials each $\pm$ 1 standard deviation.

\textbf{MNIST - fully connected layer - 1 hidden layer:}
We train a network with a fully connected architecture and 1 hidden layer on MNIST. 

We chose EP hyperparameters ($T$, $K$, $\beta$) close to those used for training the same architecture but with binary synapses and full-precision activations. At the same time we tuned BOP hyperparameters in order to fit the flipping metric in the range leading to successful training as described in Section \ref{sec:bin-W-eqprop}.

We initialize one scaling factor per weight matrix with the method described in Alg. \ref{alg:init-alpha}. Simulations with a learnt scaling factors gave results only for 1 hidden layer. When we deepened the network to be trained, learning the scaling factor does not behave well, which we think it is due to the nudging strategy (notably when $\beta<0$) which does not give an accurate estimation of the gradient.

To reach an accuracy at the level of reported results in the literature with such architecture trained by EP on MNIST, we needed to increase by 16 the number of neurons in the hidden layer which gives the architecture we trained: 784-8192-100. We chose 100 output neurons as it is approximately the number of input units times the sparsity of MNIST data and 100 has shown to perform the best.

We report all hyperparameters in Table \ref{tab:hyper-bin-W-acti-EP}. We initialize the biases with the native PyTorch random initialization and the state of the neurons to one as it has proven to perform better.

\begin{figure}[ht!]
    \begin{minipage}[c]{.46\linewidth}
        \centering
        \includegraphics[width=1\textwidth]{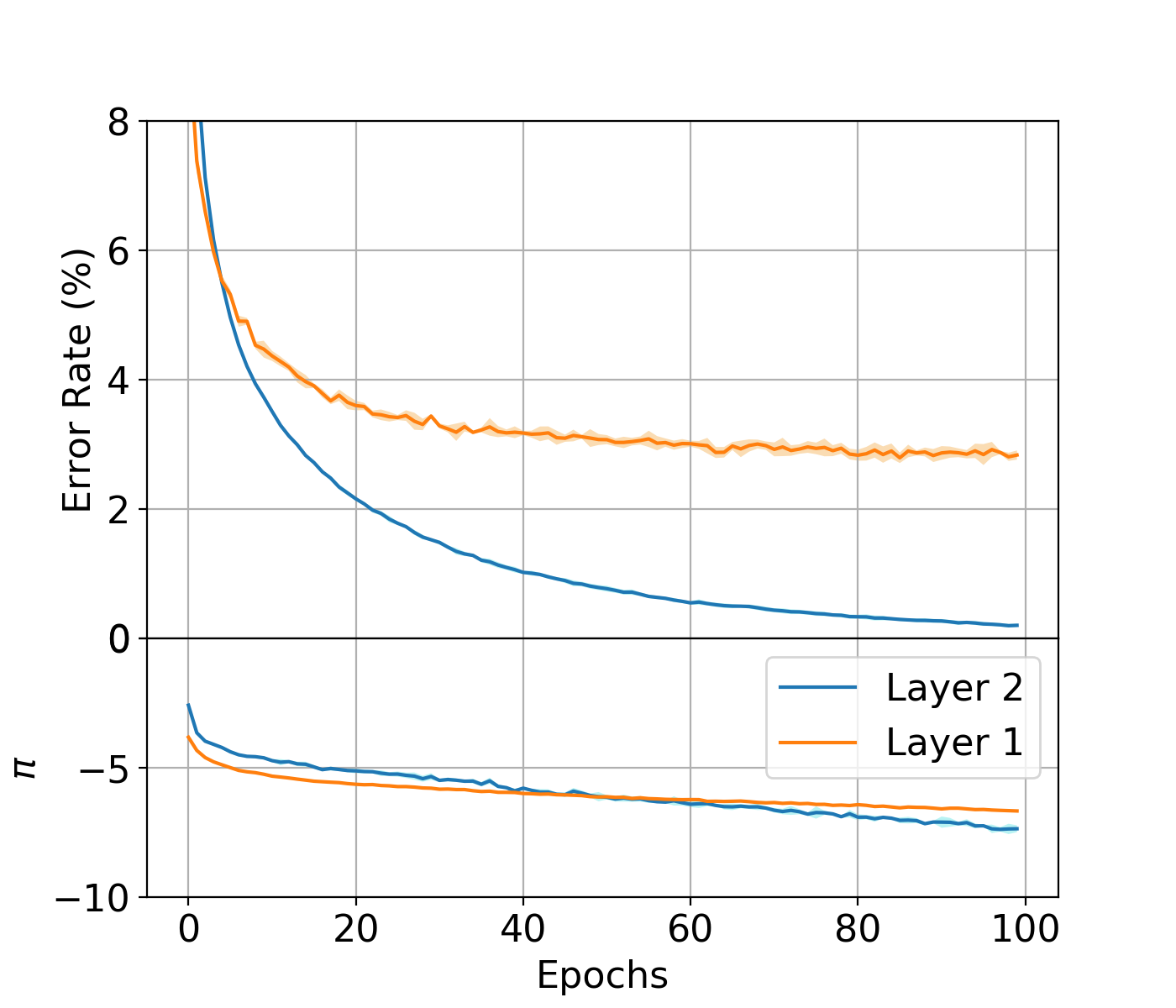}
        \caption{Top: Train (blue) and test (orange) error on MNIST with a fully connected architecture with one hidden layer of 8192 neurons trained with EP with binary synapses and binary activations - The scaling factors are fixed - Down: metric of weights flipped during an epoch, given for each weights matrix from input to output.}
        \label{fig:MNIST-1fc-8192-EP}
    \end{minipage}
    \hfill%
    \begin{minipage}[c]{.46\linewidth}
        \centering
        \includegraphics[width=1\textwidth]{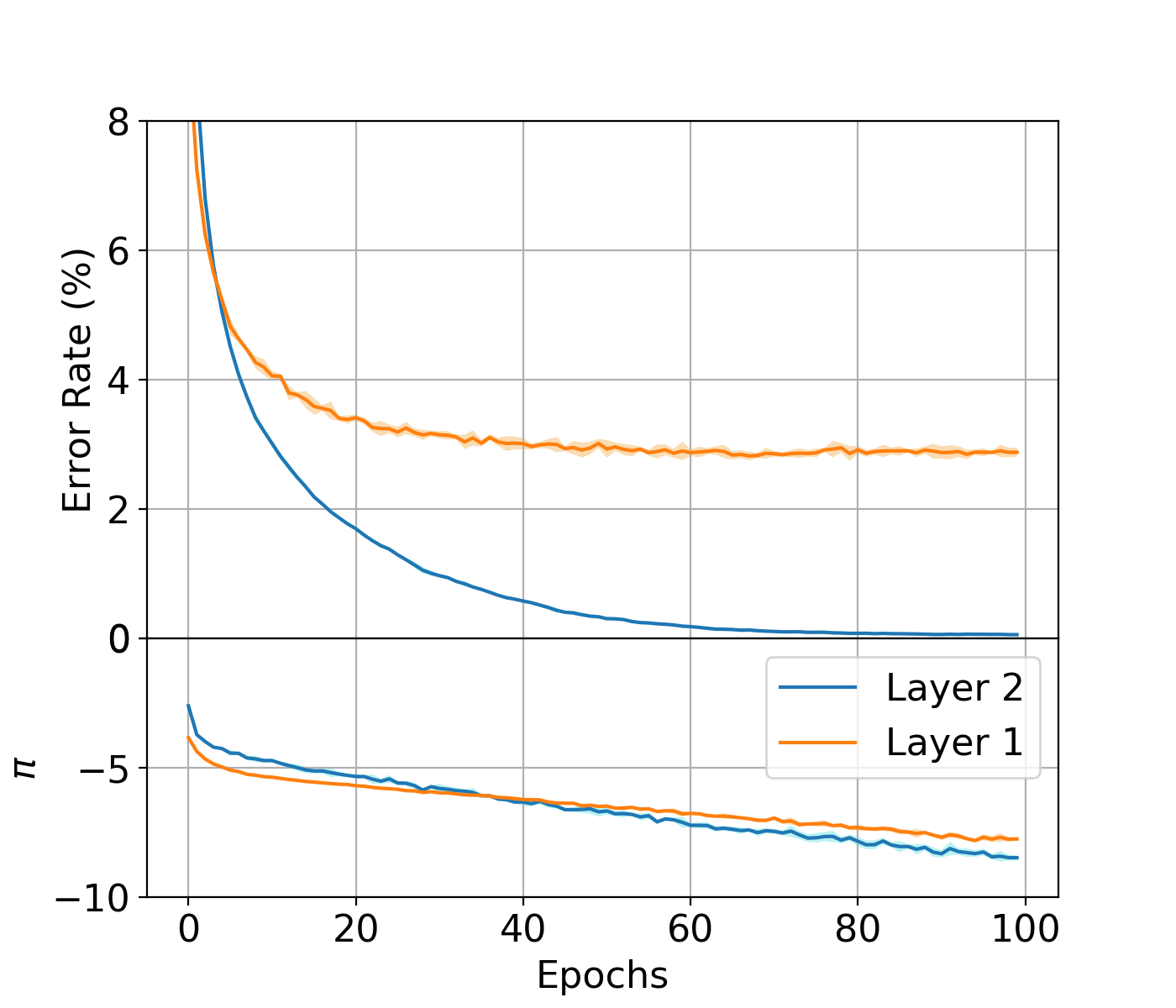}
        \caption{Top: Train (blue) and test (orange) error on MNIST with a fully connected architecture with one hidden layer of 8192 neurons trained with EP with binary synapses and binary activations - The scaling factor is learnt - Down: metric of weights flipped during an epoch, given for each weights matrix from input to output.}
        \label{fig:MNIST-1fc-8192-EP-learnAlpha}
    \end{minipage}
\end{figure}

We performed two sets of simulations: 
\begin{itemize}
    \item Simulations where the scaling factors were fixed which achieve accuracy (Table \ref{tab:results-table-bin-EP}, Fig. \ref{fig:MNIST-1fc-8192-EP}) close to those reported in the EP literature: \cite{NIPS2019_8930}.
    \item Simulations where the scaling factors were learnt. We show that learning the scaling factors only improve a little bit the training error: -0.1\% but not the testing  error: Fig. \ref{fig:MNIST-1fc-8192-EP}, Fig. \ref{fig:MNIST-1fc-8192-EP-learnAlpha}.
\end{itemize}

\textbf{MNIST - fully connected layer - 2 hidden layers}
We train a network with a fully connected architecture and 1 hidden layer on MNIST.

We chose EP hyperparameters (T,K,$\beta$) close to those used for training the same architecture but with binary synapses and full-precision activations. At the same time we tuned BOP hyperparameters in order to fit the flipping metric in the range leading to successful trainings as described in Section \ref{sec:bin-eqprop}.

We initialize one scaling factor per weight matrix with the method described in Alg. \ref{alg:init-alpha}. 

We chose 6000 output neurons as it gives the best accuracy but also as it scales as the number of hidden neurons in the penultimate hidden layer times some sparsity in the layer.

We initially perform a nudging with the sign of $\beta$ chosen randomly at each mini-batch. But when we nudge the system with $\beta<0$, it appears that we should let the system evolve during K time steps with K very large (of the order of at least 500 time steps). Finally, we chose to nudge only using the sign of $\beta>0$ despite the trainings perform less than if we used the sign of beta randomly. Monitoring the temporal evolution of some neurons in the network can also help at tuning EP hyperparameters.

To reach an accuracy at levels of reported results in the literature with such architecture trained by EP on MNIST, we used the following architecture we trained: 784-8192-8192-6000. We report all hyperparameters in Table \ref{tab:hyper-bin-W-acti-EP}. We initialize the biases with the native PyTorch random initialization and the state of the neurons to one as it has proven to perform better.

We report all hyperparameters in Table \ref{tab:hyper-bin-W-acti-EP}. We initialize the biases with the native PyTorch random initialization and the state of the neurons to one as it has proven to perform better.

\begin{figure}[ht!]
    \centering
    \includegraphics[width=0.5\textwidth]{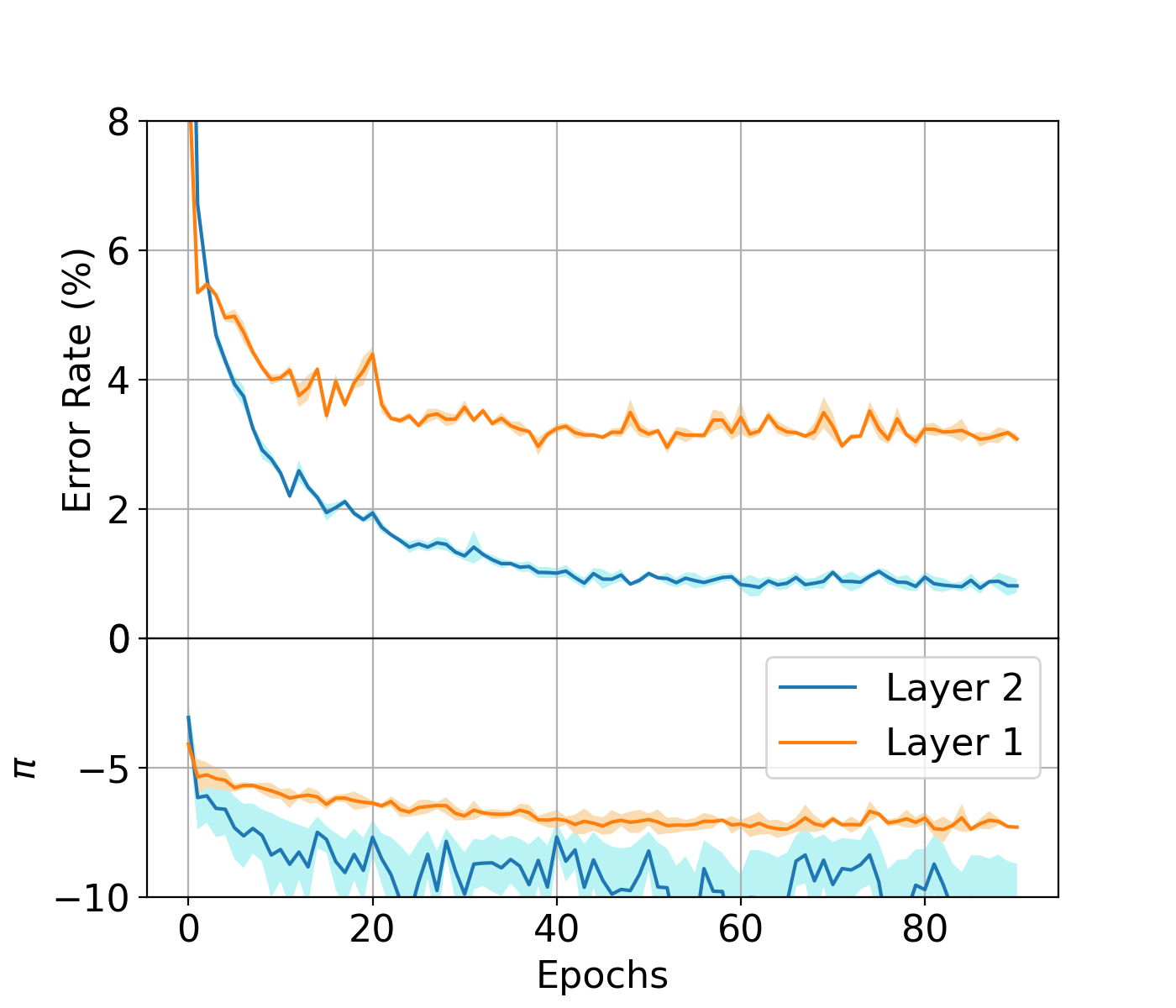}
    \caption{Top: Train (blue) and test (orange) error on MNIST with a fully connected architecture with two hidden layers of 8192 neurons trained with EP with binary synapses \& binary neurons - The scaling factors are fixed - Down: metric of weights flipped during an epoch, given for each weights matrix from input to output.}
  \label{fig:MNIST-2fc-8192-8192-full-bin-EP}
\end{figure}

\textbf{MNIST - convolutional architecture}
We train a convolutional network on MNIST. The architecture used consists in the following: 2 convolutional layers of respectively 256 and 512 channels. We use convolutional kernels of size $5\times 5$, padding of 1 and a stride of 1. Each convolutional operation is followed by a $3\time3$ Max Pooling operation with a stride of $3$. We flatten the output of the last convolutional layer to feed the output layer of 700 neurons.

We tuned BOP hyperparameters to make the metric in the range below -5.

The scaling factors $\alpha$ are initialized channel-wise in each convolutional layer which gives 256 scaling factors for the first convolutional layer and 512 scaling factors for the second convolutional layer with the architecture used here. 

Again, learning the scaling factors did not show better accuracy and could be also linked to the nudging strategy.

We initialize the biases at 0
and the state of the neurons to one as it has proven to perform better.

Finally, here we adopted another nudging implementation: although the nudging is usually performed by adding the derivative of the loss function with respect to the units of the output layer $+\beta(\text{y}-\hat{y})$, we implemented a constant nudge: $+\beta(\text{y}-\hat{y}_{*})$, where $\hat{y}_{*}$ stands for the first steady state reached by the output units at the end of the first phase. This nudge has shown to perform better than the classic nudge.
 
We report all hyperparameters in Table \ref{tab:hyper-bin-W-acti-EP}. We initialize the biases with the native PyTorch random initialization and the state of the neurons to one as it has proven to perform better.

\begin{figure}[ht!]
    \centering
    \includegraphics[width=0.5\textwidth]{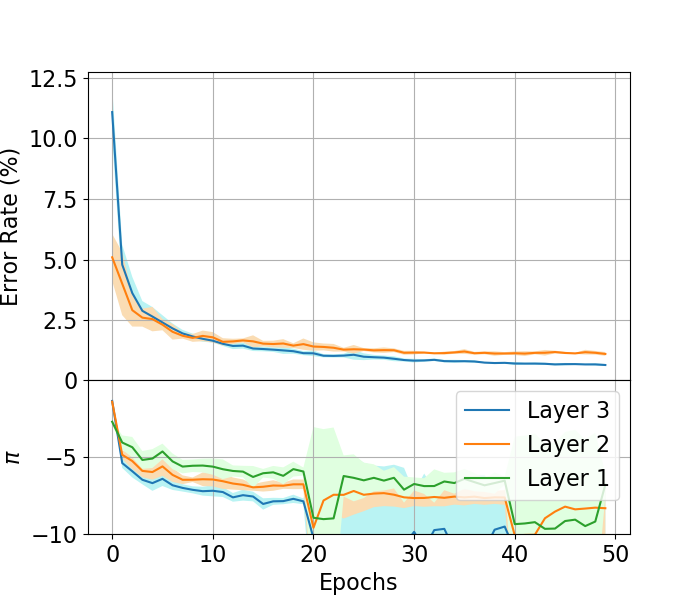}
    \caption{Top: Train (blue) and test (orange) error on MNIST with a convolutional architecture with 2 convolutional layers of respectively 256 and 512 channels trained with EP with binary synapses \& binary neurons - The scaling factors are fixed - Down: metric of weights flipped during an epoch, given for each weights matrix from input to output.}
  \label{fig:MNIST-conv-bin-EP}
\end{figure}

\begin{table*}[!ht]
 \caption{Hyperparameters used for training systems with EP and binary synapses with binary neurons - $\gamma$ is layer-dependent and given from input to output layer when multiple values are given - $\gamma$ has the same value for all layers when a single value is given.}
  \centering
  \begin{tabular}{lllccccccc}
  \toprule
     & & & \multicolumn{3}{c}{EP} & & \multicolumn{2}{c}{BOP} &  \\
    \cline{4-6}
    \cline{8-9}
    Dataset     & Method     & Architecture  & T & K & $\beta$ & & $\gamma$ & $\tau$ & $lrBias$\\
    \midrule
    MNIST & fc  & 784-8192-100 & 20 & 10 & 2 & & 2e-6 & 2.5e-7 - 2e-7 & 1e-7   \\
    MNIST     & fc & 784-8192(2)-8000 & 30 & 80 & 2 & & 1e-6 & 2e-8 - 1e-8 - 5e-8 & 1e-6   \\
    MNIST     & conv  & 1-256-512-1600(fc) & 100  & 50  & 1 & & 5e-8 & 8e-8 - 8e-8 - 2e-7 & 2e-6 - 5e-6 - 1e-5  \\
    \bottomrule
  \end{tabular}
  \label{tab:hyper-bin-W-acti-EP}
\end{table*}

\end{document}